\documentclass[sigconf, nonacm]{acmart}

\newcommand\vldbdoi{XX.XX/XXX.XX}
\newcommand\vldbpages{XXX-XXX}
\newcommand\vldbvolume{17}
\newcommand\vldbissue{3}
\newcommand\vldbyear{2023}
\newcommand\vldbauthors{\authors}
\newcommand\vldbtitle{\shorttitle} 
\newcommand\vldbavailabilityurl{https://github.com/17000cyh/IMDiffusion.git}
\newcommand\vldbpagestyle{empty}

\usepackage{color}
\usepackage{amsmath}    
\usepackage{tabularray}
\usepackage{multirow}
\usepackage{natbib}
\usepackage{graphicx}
\usepackage{subfigure}
\usepackage{textcomp}
\usepackage{xcolor}
\usepackage{algpseudocode}
\usepackage{algorithm}

\theoremstyle{definition}

\theoremstyle{remark}

\usepackage{graphicx}
\usepackage{tabularx}
\usepackage{amsmath,xparse,mleftright}
\usepackage[export]{adjustbox}
\usepackage{balance}
\usepackage{tcolorbox}
\usepackage{bbm}
\usepackage{dsfont}

\usepackage{enumitem}
\usepackage{xcolor}
\usepackage{pifont}
\usepackage{xspace}

\usepackage{setspace}
\usepackage{url}

\newcommand{\eg}{\emph{e.g.},\xspace}
\newcommand{\ie}{\emph{i.e.},\xspace}

\newcommand{\name}{\textsc{ImDiffusion}\xspace}
\newcommand{\model}{\textsc{ImTransformer}\xspace}
\newcommand{\modelt}{\textsc{ImTransformer}}

\newcommand{\rv}[1]{\textcolor{black} {#1}}

\setcopyright{none}
\algnewcommand{\Inputs}[1]{%
  \State \textbf{Inputs:}
  \Statex \hspace*{\algorithmicindent}\parbox[t]{.8\linewidth}{\raggedright #1}
}
\algnewcommand{\Initialize}[1]{%
  \State \textbf{Initialise:}
  \Statex \hspace*{\algorithmicindent}\parbox[t]{.8\linewidth}{\raggedright #1}
}

\begin{document}
\title[\name: Imputed Diffusion Models for Multivariate Time Series Anomaly Detection]{\name: Imputed Diffusion Models for Multivariate \\Time Series Anomaly Detection}

\settopmatter{authorsperrow=4} 

\author{Yuhang Chen}
\affiliation{%
     \institution{Peking University}
     \country{}
 }
 \email{2101210553@pku.edu.cn}
 \authornote{This work was completed during their internship at Microsoft Research Asia.}

\author{Chaoyun Zhang}
\affiliation{%
\institution{Microsoft}
\country{}
 }
  \email{chaoyun.zhang@microsoft.com}
 \authornote{Corresponding author.}

\author{Minghua Ma}
\author{Yudong Liu}
\affiliation{%
\institution{Microsoft}
\country{}
 }
\email{minghuama@microsoft.com}
\email{bahuangliuhe@pku.edu.cn}

\author{Ruomeng Ding}
\affiliation{%
\institution{Georgia Institute of Technology}
\country{}
 }
 \email{rmding@gatech.edu}
 \authornotemark[1]

\author{Bowen Li}
\affiliation{%
\institution{Tsinghua University}
\country{}
 }
 \email{libowen.ne@gmail.com}

 \author{Shilin He}
\affiliation{%
     \institution{Microsoft}
     \country{}
 }
\email{shilin.he@microsoft.com}

\author{Saravan Rajmohan}
\affiliation{%
\institution{Microsoft 365}
\country{}
 }
 \email{saravar@microsoft.com}

\author{Qingwei Lin}
\author{Dongmei Zhang}
\affiliation{%
     \institution{Microsoft}
     \country{}
 }
\email{{qlin,dongmeiz}@microsoft.com}



\begin{abstract}
Anomaly detection in multivariate time series data is of paramount importance for ensuring the efficient operation of large-scale systems across diverse domains. However, accurately detecting anomalies in such data poses significant challenges due to the need for precise modeling of complex multivariate time series data. Existing approaches, including forecasting and reconstruction-based methods, struggle to address these challenges effectively. To overcome these limitations, we propose a novel anomaly detection framework named \name, which combines time series imputation and diffusion models to achieve accurate and robust anomaly detection. The imputation-based approach employed by \name leverages the information from neighboring values in the time series, enabling precise modeling of temporal and inter-correlated dependencies, reducing uncertainty in the data, thereby enhancing the robustness of the anomaly detection process.  \name further leverages diffusion models as time series imputers to accurately capture complex dependencies. We leverage the step-by-step denoised outputs generated during the inference process to serve as valuable signals for anomaly prediction, resulting in improved accuracy and robustness of the detection process.

We evaluate the performance of \name via extensive experiments on benchmark datasets. The results demonstrate that our proposed framework significantly outperforms state-of-the-art approaches in terms of detection accuracy and timeliness. \name is further integrated into the real production system in Microsoft and observes a remarkable 11.4\% increase in detection F1 score compared to the legacy approach. To the best of our knowledge, \name represents a pioneering approach that combines imputation-based techniques with time series anomaly detection, while introducing the novel use of diffusion models to the field.

\end{abstract}



\maketitle

\pagestyle{\vldbpagestyle}
\begingroup\small\noindent\raggedright\textbf{PVLDB Reference Format:}\\
\vldbauthors. \vldbtitle. PVLDB, \vldbvolume(\vldbissue): \vldbpages, \vldbyear.\\
\href{https://doi.org/\vldbdoi}{doi:\vldbdoi}
\endgroup
\begingroup
\renewcommand\thefootnote{}\footnote{\noindent
This work is licensed under the Creative Commons BY-NC-ND 4.0 International License. Visit \url{https://creativecommons.org/licenses/by-nc-nd/4.0/} to view a copy of this license. For any use beyond those covered by this license, obtain permission by emailing \href{mailto:info@vldb.org}{info@vldb.org}. Copyright is held by the owner/author(s). Publication rights licensed to the VLDB Endowment. \\
\raggedright Proceedings of the VLDB Endowment, Vol. \vldbvolume, No. \vldbissue\ %
ISSN 2150-8097. \\
\href{https://doi.org/\vldbdoi}{doi:\vldbdoi} \\
}\addtocounter{footnote}{-1}\endgroup

\ifdefempty{\vldbavailabilityurl}{}{
\vspace{.3cm}
\begingroup\small\noindent\raggedright\textbf{PVLDB Artifact Availability:}\\
The source code, data, and/or other artifacts have been made available at \url{\vldbavailabilityurl}.
\endgroup
}

\section{Introduction}
The efficient operation of large-scale systems or entities heavily relies on the generation and analysis of extensive and high-dimensional time series data. These data serve as a vital source of information for continuous monitoring and ensuring the optimal functioning of these systems. However, within these systems, various abnormal events may occur, resulting in deviations from the expected downstream performance of numerous applications \cite{blazquez2021review, schmidl2022anomaly,  jin2023assess}. These anomalous events can encompass a broad spectrum of issues, including production faults \cite{chen2023empowering, ma2022empirical}, delivery bottlenecks \cite{ibidunmoye2015performance}, system defects \cite{wenig2022timeeval, yan2023aegis}, or irregular heart rhythms \cite{li2020survey}. When different time series dimensions are combined, they form a multivariate time series (MTS). The detection of anomalies in MTS data has emerged as a critical task across diverse domains. Industries spanning manufacturing, finance, and healthcare monitoring, have recognized the importance of anomaly detection in maintaining operational efficiency and minimizing disruptions \cite{schmidl2022anomaly, jacob2021exathlon}, and the field of MTS anomaly detection has garnered significant attention from both academia and industry \cite{boniol2021sand, boniol2020graphan, lu2022matrix, boniol2020automated, alnegheimish2022sintel}.

However, achieving accurate anomaly detection on MTS data is not straightforward, as it necessitates precise modeling of time series data \cite{blazquez2021review, zeng2023traceark, ma2018robust}. The complexity of modern large-scale systems introduces additional challenges, as their performance is monitored by multiple sensors, generating heterogeneous time series data that encompasses multidimensional, intricate, and interrelated temporal information \cite{li2021multivariate, ma2021jump}. Modeling complex correlations like these requires a high level of capability from the model.
Furthermore, time series data often displays significant variability \cite{ma2020diagnosing}, leading to increased levels of uncertainty. This variability can sometimes result in erroneous identification of anomalies. This adds complexity to the anomaly detection process, as the detector must effectively differentiate between stochastic anomalies and other variations to achieve robust detection performance \cite{su2019robust, zhao2023robust}.


The aforementioned challenges have spurred the emergence of numerous self-supervised learning solutions aimed at automating anomaly detection. Recent methods can be classified into various categories \cite{schmidl2022anomaly}, where forecasting \cite{zhao2020multivariate, munir2018deepant} and reconstruction-based \cite{zhang2019deep,ma2021jump, tuli2022tranad} approaches have been most widely employed. The former leverages past information to predict future values in the time series and utilizes the prediction error as an indicator for anomaly detection. However, future time series values can exhibit high levels of uncertainty and variability, making them inherently challenging to accurately predict in complex real-world systems. Relying solely on forecasting-based methods may have a detrimental impact on anomaly detection performance \cite{li2021multivariate, malhotra2016lstm}. On the other hand, reconstruction-based methods encode entire time sequences into an embedding space. Anomaly labels are then inferred based on the reconstruction error. Since these approaches operate and need to reconstruct the entire time series, their performance heavily relies on the capabilities of the reconstruction model \cite{ma2021jump}. In cases where the original data exhibit heterogeneity, complexity, and inter-dependencies, reconstruction-based methods may encounter challenges in achieving low overall reconstruction error and variance \cite{li2019mad, audibert2020usad}. As a result, the anomaly detection performance of such approaches may be sub-optimal. Given these considerations, there is a clear need to rethink and enhance forecasting and reconstruction approaches to achieve accurate and robust anomaly detection.

\begin{figure}[t]
\centering
\includegraphics[width=0.95\columnwidth]{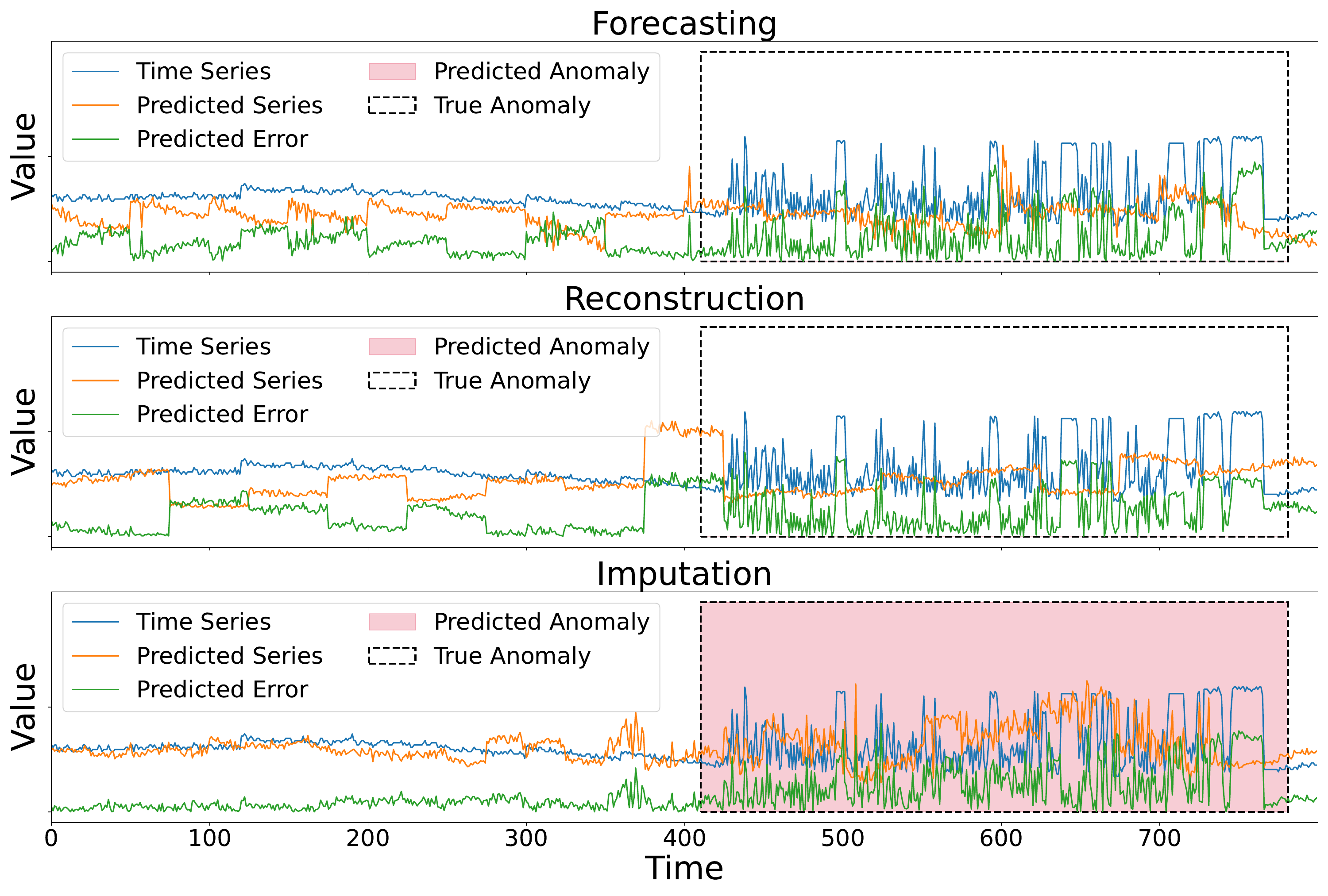}
\vspace*{-1.5em}
\caption{\rv{Examples of reconstruction, forecasting and imputation modeling of time series for anomaly detection.}
\label{fig:pretask_case}}
\vspace*{-2em}
\end{figure}

To address these challenges and overcome the limitations of existing approaches, we propose a novel anomaly detector named \name. This detector combines the use of time series imputation \cite{fang2020time} and diffusion models \cite{ho2020denoising} to achieve accurate and robust anomaly detection. \name employs dedicated grating data masking into the time series data, creating unobserved data points. It then utilizes diffusion models to accurately model the MTS and impute the missing values caused by the data masking. The imputation error is subsequently used as an indicator to determine the anomalies. The imputation-based approach employed by \name offers distinct advantages over forecasting and reconstruction methods. Firstly, it leverages neighboring values in the time series as additional conditional information, enabling a more accurate modeling of the temporal and inter-correlated dependencies present in MTS. Secondly, the reference information from neighboring values helps to reduce uncertainty in predictions, and thereby enhancing the robustness of the detection process. Fig.~\ref{fig:pretask_case} presents an example in which forecasting, reconstruction, and imputation methods are employed to predict a time series using diffusion models. 
\rv{The forecasting method employs a 50-step MTS for observation and predicts the subsequent 50-step MTS. The reconstruction method recovers the entire 100-step MTS. Meanwhile, the imputation method is carried out using the grating data masking. Observe that while all approaches yield comparable errors during the outlier period, the imputation approach achieves a lower error within the normal range due to its superior MTS modeling ability. This attribute enables it to establish a more distinct decision boundary for anomaly identification. As a result, only the imputation method successfully identifies the period of anomaly.} 
We therefore employ time series imputation for accurate self-supervised modeling of time series, which forms the foundation of our proposed \name.

To enhance the performance of anomaly detection, \name leverages the exceptional unsupervised modeling capability of diffusion models \cite{ho2020denoising} for imputation. Diffusion models have demonstrated superior performance in unsupervised image generation, surpassing traditional generative models such as GANs \cite{goodfellow2020generative} and VAEs \cite{kingma2013auto}. They have also been successfully applied to model complex temporal and inter-metric dependencies in MTS, showcasing remarkable abilities in forecasting \cite{rasul2021autoregressive} and imputation \cite{tashiro2021csdi}. We employ a dedicated diffusion model as the time series imputer, replacing traditional forecasting and reconstruction models. This brings several advantages to anomaly detection, namely \emph{(i)} it enables better modeling of complex correlations within MTS data; \emph{(ii)} it allows for stochastic modeling of time series through the noise/denoising processes involved in the imputation; \emph{(iii)} the step-by-step outputs generated during the imputation inference serve as additional signals for determining the anomaly labels in an ensemble manner. These unique advantages of diffusion models enable precise capturing of the complex dependencies and inherent stochasticity present in time series data, and further enhance the robustness of anomaly detection through ensembling techniques.

By integrating imputation and diffusion models, our proposed \name achieves exceptional accuracy and timeliness in anomaly detection for both offline and online evaluation in real production. Overall, this paper presents the following contributions:
\begin{itemize}[leftmargin=*]
    \item We introduce \name, a novel framework based on the imputed diffusion model, which accurately captures the inherent dependency and stochasticity of MTS data, leading to precise and robust anomaly detection.
    \item We develop a grating masking strategy to create missing values in the data for imputation. This strategy enhances the decision boundary between normal and abnormal data, resulting in improved anomaly detection performance.
    \item \name leverages the step-by-step denoised outputs of the diffusion model's unique inference process as additional signals for anomaly prediction in an ensemble voting  manner. This approach further enhances inference accuracy and robustness.
    \item We conduct extensive experiments comparing \name with 10 state-of-the-art anomaly detection baselines on 6 datasets. Results show that \name significantly outperforms other approaches in terms of both detection accuracy and timeliness.
    \item We integrate \name in the real production of the Microsoft email delivery microservice system. The framework exhibits an 11.4\% higher detection accuracy compared to the legacy online approach, which significantly improves the system's reliability.
\end{itemize}
To the best of our knowledge, \name is the pioneering approach that combines imputation-based techniques with MTS anomaly detection, and it \rv{pushes the methodology boundaries} by first applying diffusion models to this field.

\vspace*{-0.5em}
\section{Related Work}

We provides an overview of relevant research in time series anomaly detection \cite{schmidl2022anomaly} and diffusion models \cite{lin2023diffusion} in this section. 

\vspace*{-0.5em}
\subsection{Time Series Anomaly Detection}
Time series anomaly detection is an important problem that has received significant attention from both the industrial and research communities \cite{zhang2019deep2, schmidl2022anomaly, paparrizos2022tsb, paparrizos2022volume, ting2022new, wenig2022timeeval, boniol2022theseus, sylligardos2023choose}. Approaches for this area can be categorized into five main classes based on the underlying detection method \cite{schmidl2022anomaly}. These categories include: \emph{(i)} forecasting methods (\eg \cite{zhao2020multivariate, munir2018deepant}), which predict future values to identify anomalies; \emph{(ii)} reconstruction methods (\eg \cite{zhang2019deep, tuli2022tranad, campos2022unsupervised}), which reconstruct the time series and identify anomalies based on the reconstruction error; \emph{(iii)} encoding methods (\eg \cite{boniol13series2graph}), which encode the time series into a different representation and detect anomalies using this encoding; \emph{(iv)} distance methods (\eg \cite{breunig2000lof, boniol2021sand}), which measure the dissimilarity between time series and identify anomalies based on the distance; \emph{(v)} distribution methods (\eg \cite{goldstein2012histogram, hochenbaum2017automatic}), which model the distribution of the time series data and detect anomalies based on deviations from the expected distribution; and \emph{(vi)} isolation tree methods (\eg \cite{liu2008isolation, cheng2019outlier}), which use tree-based structures to isolate anomalies.

Among the various approaches explored in the literature, forecasting and reconstruction methods have gained significant popularity due to their reported effectiveness. For instance, Omnianomaly \cite{su2019robust} employs a combination of GRU and VAE to learn robust representations of time series. It also utilizes the Peaks-Over-Threshold (POT) method to dynamically select appropriate thresholds for anomaly detection. MTAD-GAT \cite{zhao2020multivariate} incorporates a graph-attention network to capture both feature and temporal correlations within time series data. By combining forecasting and reconstruction models, it achieves improved anomaly detection performance. MAD-GAN \cite{li2019mad} takes advantage of the discriminator's loss in a GAN as an additional indicator for detecting anomalies. More recently, TranAD \cite{tuli2022tranad} introduces attention mechanisms in transformer models and incorporates adversarial training to jointly enhance the accuracy of anomaly detection.

The \name introduced in this paper distinguishes from previous approaches by employing time series imputation to improve time series modeling and enhance anomaly detection performance. 

\vspace*{-0.5em}
\subsection{Diffusion Model}
Recently, diffusion models \cite{yang2022diffusion, lin2023diffusion} have garnered increasing attention in the field of AI generated content \cite{ramesh2022hierarchical, rombach2022high}. While their potential in the domain of time series modeling and anomaly detection is relatively new, researchers have begun to explore their application in these areas. For instance, CSDI \cite{tashiro2021csdi} utilizes a probabilistic diffusion model for time series imputation, outperforming deterministic baselines. TimeGrad \cite{rasul2021autoregressive} applies diffusion models in an autoregressive manner to generate future time sequences for forecasting. This approach achieves good performance in extrapolating into the future while maintaining computational tractability. Additionally, diffusion models have been employed in time series generation. In \cite{lim2023regular}, diffusion models are used as score-based generative models to synthesize time-series data, resulting in superior generation quality and diversity compared to baseline approaches.

Diffusion models have also been explored for image anomaly detection. In \cite{wolleb2022diffusion}, denoising diffusion implicit models \cite{songdenoising} are combined with classifier guidance to identify anomalous regions in medical images. This produces highly detailed anomaly maps without the need for a complex training procedure. Similarly, in \cite{pinaya2022fast}, diffusion models are used to eliminate bias and mitigate accumulated prediction errors, thereby enhancing anomaly segmentation in CT data. The DiffusionAD \cite{zhang2023diffusionad} formulates anomaly detection as a ``noise-to-norm'' paradigm, requiring only one diffusion reverse process step to achieve satisfactory performance in image anomaly detection. This significantly improves the inference efficiency.

To the best of our knowledge, our proposed \name represents the first utilization of diffusion models specifically tailored for the task of MTS anomaly detection.

\vspace*{-0.5em}
\section{Preliminary}
In this section, we present the problem of MTS anomaly detection and time series imputation, and an overview of diffusion models.

\vspace*{-0.5em}
\subsection{Multivariate Time Series Anomaly Detection} 
We consider a collection of MTS denoted as $\mathcal{X}$, which encompasses measurements recorded from timestamp $1$ to $L$. Specifically:
\begin{align}
\mathcal{X} = \{ \mathbf{x}_1, \mathbf{x}_2, \cdots \mathbf{x}_L\},
\end{align}
where $\mathbf{x}_l \in \mathbb{R}^K$ represents an $K$-dimensional vector at time $l$, \ie $\mathbf{x}_l = \{ x_l^1, x_l^2, \cdots x_l^K\}$. The objective of MTS anomaly detection is to determine whether an observation $\mathbf{x}_l$ is anomalous or not. By employing $y_l \in \{0 ,1\}$ to indicate the presence of an anomaly (with 0 denoting no anomaly and 1 denoting an anomaly), the goal transforms into predicting a sequence of anomaly labels for each timestamp, namely $Y = \{y_1, y_2, \cdots, y_L\}$. 

\vspace*{-0.5em}
\subsection{Time Series Imputation}
\name leverages the prediction error resulting from the imputation  \cite{tashiro2021csdi} of intentionally masked values within a time series to infer the anomaly labels. The mask is denoted as $\mathcal{M} = \{ m_{l \in 1:L, k \in 1:K} \} \in \{0, 1\}$, where $m = 1$ indicates that $x_l^k$ is observed, while $0$ signifies that it is missing. The mask $\mathcal{M}$ possesses the same dimensionality as the time series $\mathcal{X}$, \ie $\mathcal{M} \in \mathbb{R}^{T\times K}$. The application of the mask $\mathcal{M}$ to the original time series $\mathcal{X}$ yields a new partially observed  time series $\mathcal{X}^\mathcal{M}$, which can be expressed as:
\begin{align}
    \mathcal{X}^\mathcal{M}  = \mathcal{X} \odot \mathcal{M}.
\end{align}
Here, the symbol $\odot$ represents the Hadamard product. Let $\mathcal{X}^{\mathcal{M}_0}$ represent the masked value where $m_{l,k} = 0$, and  $\mathcal{X}^{\mathcal{M}_1}$ represent the observed values where $m_{l,k} = 1$, the objective of the imputation process is to estimate the missing values in $\mathcal{X}^\mathcal{M}$, \ie $p(\mathcal{X}^{\mathcal{M}_0}\mid\mathcal{X}^{\mathcal{M}_1})$. Several tasks, such as interpolation  \cite{lepot2017interpolation, shuklainterpolation, jhin2022exit} and forecasting \cite{zhang2021cloudlstm, zhang2019multi, zhou2021informer}, can be considered as instances of time series imputation.

\vspace*{-0.5em}
\subsection{Denoising Diffusion Model}
Our \name is based on the diffusion models \cite{sohl2015deep}, a well-known generative model that draws inspiration from non-equilibrium thermodynamics. Diffusion models follow a two-step process for data generation. Firstly, it introduces noise to the input incrementally, akin to a forward process. Secondly, it learns to generate new samples by progressively removing the noise from a sample noise vector, thereby resembling a reverse process.
During the forward process, Gaussian noise is incrementally added to the initial input sample $\mathcal{X}_0$ over $T$ steps. Mathematically, this can be represented as $q(\mathcal{X}_{1:T}\mid \mathcal{X}_0) := \prod_{t=1}^{T} q(\mathcal{X}_{t}\mid\mathcal{X}_{t-1}),$
where
\begin{align}\label{eq:forward}
q(\mathcal{X}_{t}\mid\mathcal{X}_{t-1}) := \mathcal{N}(\mathcal{X}_{t};\sqrt{1-\beta_t}\mathcal{X}_{t-1}, \beta_t\mathbf{I}).  
\end{align}
Here, $\beta$ is a positive noise level constant that can either be learned or predefined. The forward process is parameterized as a Markov chain, as the values of $\mathcal{X}_{t}$  solely depend on $\mathcal{X}_{t-1}$. The final step, $\mathcal{X}_{T}$, is fully corrupted and becomes random noise. Its distribution can be expressed in closed form as $q(\mathcal{X}_{T} \mid \mathcal{X}_{0}) = \mathcal{N}(\mathcal{X}_{T};\sqrt{\alpha_t}\mathcal{X}_{0}, (1-\alpha_t)\mathbf{I})$, where $\Tilde{\alpha}_t := 1 - \beta_t$ and  $\alpha_t := \prod_{i=1}^t \Tilde{\alpha}_i$. Next, $\mathcal{X}_{T}$ can be represented as $\mathcal{X}_{T} = \sqrt{\alpha_T}\mathcal{X}_{0} + (1-\alpha_T)\epsilon $, where $\epsilon \sim \mathcal{N}(\mathbf{0},\mathbf{I})$. 

Conversely, we employ a machine learning model with learnable parameters $\Theta$ to denoise $\mathcal{X}_{T}$ and reconstruct $\mathcal{X}_{0}$. This is accomplished by iteratively computing the following Gaussian transitions:
\begin{align}\label{eq:reverse}
    p_{\Theta} (\mathcal{X}_{t-1}\mid\mathcal{X}_{t}) := \mathcal{N}(\mathcal{X}_{t-1}; \mathbf{\mu}_\Theta(\mathcal{X}_{t}, t), \mathbf{\Sigma}_\Theta(\mathcal{X}_{t}, t)\mathbf{I}).
\end{align}
As a result, the joint distribution can be expressed as $p_\Theta(\mathcal{X}_{0:T}) = p(\mathcal{X}_{T})\prod_{t=1}^T p_{\Theta} (\mathcal{X}_{t-1}\mid\mathcal{X}_{t})$.

The Denoising Diffusion Probabilistic Model (DDPM) \cite{ho2020denoising} simplifies the reverse process by adopting a fixed variance, result in:
\begin{align}\label{eq:ddpm}
\mathbf{\mu}_\Theta(\mathcal{X}_{t}, t) := \frac{1}{\alpha_t} \left (\mathcal{X}_{t} - \frac{\beta_t}{\sqrt{1-\alpha_t}} \epsilon_\Theta(\mathcal{X}_{t}, t) \right), \ \ \mathbf{\Sigma}_\Theta(\mathcal{X}_{t}, t) = \sqrt{\Tilde{\beta}_t}.
\end{align}
Here $\Tilde{\beta}_t = \left\{\begin{matrix}
 \frac{1-\alpha_{t-1}}{1-\alpha_t}\beta_t, &t>1\\
\beta_1, &t=1
\end{matrix}\right.$, and $\epsilon_\Theta$ represents a trainable denoising function. By employing Jensen's inequality and the speeding-up parameterization from DDPM, the reverse process can be solved by training a model to optimize the following objective function:
\begin{align}
    \min_\Theta \mathcal{L}(\Theta) := \min_\Theta \mathbb{E}_{\mathcal{X}_0\sim q(\mathcal{X}_0), \epsilon\sim\mathcal{N}(\mathbf{0}, \mathbf{I}),t}\mid\mid\epsilon -\epsilon _\Theta(\mathcal{X}_t,t)\mid\mid^2,
\end{align}
where $\mathcal{X}_t = \sqrt{\alpha_t}\mathcal{X}_0 + (1-\alpha_t)\epsilon$, \rv{$\mathcal{X}_{0}$ is the complete data sample that is unaffected by diffusion process noise, and $q(\mathcal{X}_0)$ is its distribution \cite{ho2020denoising}.} 
The denoising function $\epsilon_\Theta$ is responsible for estimating the noise added to the corrupted input $\mathcal{X}_t$. Once trained, given an arbitrary noise vector, we can generate a new sample by progressively denoising using  $\mathcal{X}_t$ and obtain a final complete sample.

\section{The Design of \name}


\name relies on time series imputation and utilizes the imputed error as a signal for anomaly detection. The imputation process is carried out in a self-supervised learning manner, where we intentionally introduce masks to the MTS, creating missing values that need to be imputed. We then train a diffusion model using \model designed specifically for imputation and subsequent anomaly detection tasks. During the inference phase, we leverage the intermediate output of the \model at different denoising steps $t$ as additional information to collectively determine the anomaly label. This ensemble approach enhances the accuracy and robustness of \name, further improving its performance.

\vspace*{-0.5em}
\subsection{Imputed Diffusion Models\label{sec:impdiffusion}}
Time series anomaly detection often relies on the construction of prediction models that accurately capture the distribution of normal data. These models are expected to exhibit higher prediction errors when anomalies occur, thereby serving as indicators and providing a decision boundary for detecting anomalies. Two commonly used types of prediction models for anomaly detection are \emph{(i)} reconstruction models, which encode the entire time series into a representation that can be reconstructed using a decoder; \emph{(ii)} forecasting models, which aim to predict future values of the time series based on historical observations \cite{hundman2018detecting}.
However, both types of prediction models have their limitations in terms of their capacity for time series modeling. \rv{When applying the diffusion model, the reconstruction method involves corrupting the entire MTS into a complete noise vector for reconstruction. However, this introduces a significant level of uncertainty, particularly when there is a lack of conditional information.} Similarly, forecasting models face challenges in accurately predicting future values, especially in the presence of anomalies, further contributing to the uncertainty.

To overcome the limitations of traditional reconstruction \cite{zhang2019deep, su2019robust, malhotra2016lstm} and forecasting models \cite{kim2018deepnap, munir2018deepant, zhao2020multivariate}, we propose the use of time series imputation as the underlying prediction model for anomaly detection in \name. We further enhance the imputation capacity of the model by incorporating state-of-the-art diffusion models. This approach offers several advantages. Firstly, it enables enhanced estimation of the data distribution by leveraging the availability of unmasked data values. This leads to improved understanding of the underlying data distribution. Secondly, the imputation-based prediction process stabilizes the inference of the diffusion model, resulting in reduced variance in its predictions. This increased stability enhances the reliability of the model's predictions. Lastly, incorporating imputation-based prediction improves the overall accuracy and robustness of subsequent anomaly detection. By combining diffusion models for time series imputation, \name achieves accurate modeling of time series data, resulting in superior performance in anomaly detection tasks.


We begin by introducing the use of score-based diffusion models for MTS data imputation \cite{tashiro2021csdi}. There are two main categories of diffusion models employed for time series imputation, distinguished by the type of input information they utilized as follows:
\begin{itemize}[leftmargin=*]
    \item \textbf{Conditioned Diffusion Models}: These models estimate the masked values conditioned on the observed data, specifically $p(\mathcal{X}^{\mathcal{M}_0}\mid\mathcal{X}^{\mathcal{M}_1})$. In this case, the observed values $\mathcal{X}^{\mathcal{M}_1}$ are not corrupted by noise and are directly provided as input in the reverse process.
    \item \textbf{Unconditional Diffusion Models}: For unconditional imputed diffusion models introduced in \cite{tashiro2021csdi}, both masked and unmasked values are corrupted by noise in the forward process. Instead of directly providing the observed data, it retains the ground-truth noise added to the unmasked values as reference inputs. This leads to the estimation of $p(\mathcal{X}^{\mathcal{M}_0}\mid\epsilon^{\mathcal{M}_1}_{1:T})$, where $\epsilon^{\mathcal{M}_1}_{1:T}$ represents the noise sequence added to the unmasked values $\mathcal{X}^{\mathcal{M}_1}$ during the forward process.
\end{itemize}

Conditional diffusion models generally outperform unconditional diffusion models in the task of imputation, resulting in lower overall prediction errors \cite{tashiro2021csdi}. This is because conditional models benefit from the direct inclusion of ground-truth unmasked data as input, which serves as reliable references for neighboring values. However, it is important to recognize the distinction between the objectives of imputation and anomaly detection. While imputation aims to minimize the error between predictions and ground truth for all data points, anomaly detection requires a clear boundary between normal and abnormal points, achieved by minimizing imputation errors only for normal data and maximizing errors for anomaly points. During the inference phase, when anomaly points happen to be unmasked and used as inputs for the prediction model, the prediction error for neighboring anomaly points is also reduced. Consequently, the prediction error becomes indistinguishable between normal and abnormal points, compromising the effectiveness of subsequent anomaly detection. The existence of unmasked anomaly points during inference blurs the clear boundary in the prediction error that is vital for accurate anomaly detection.

\begin{figure}[t]
\centering
\includegraphics[width=0.95\columnwidth]{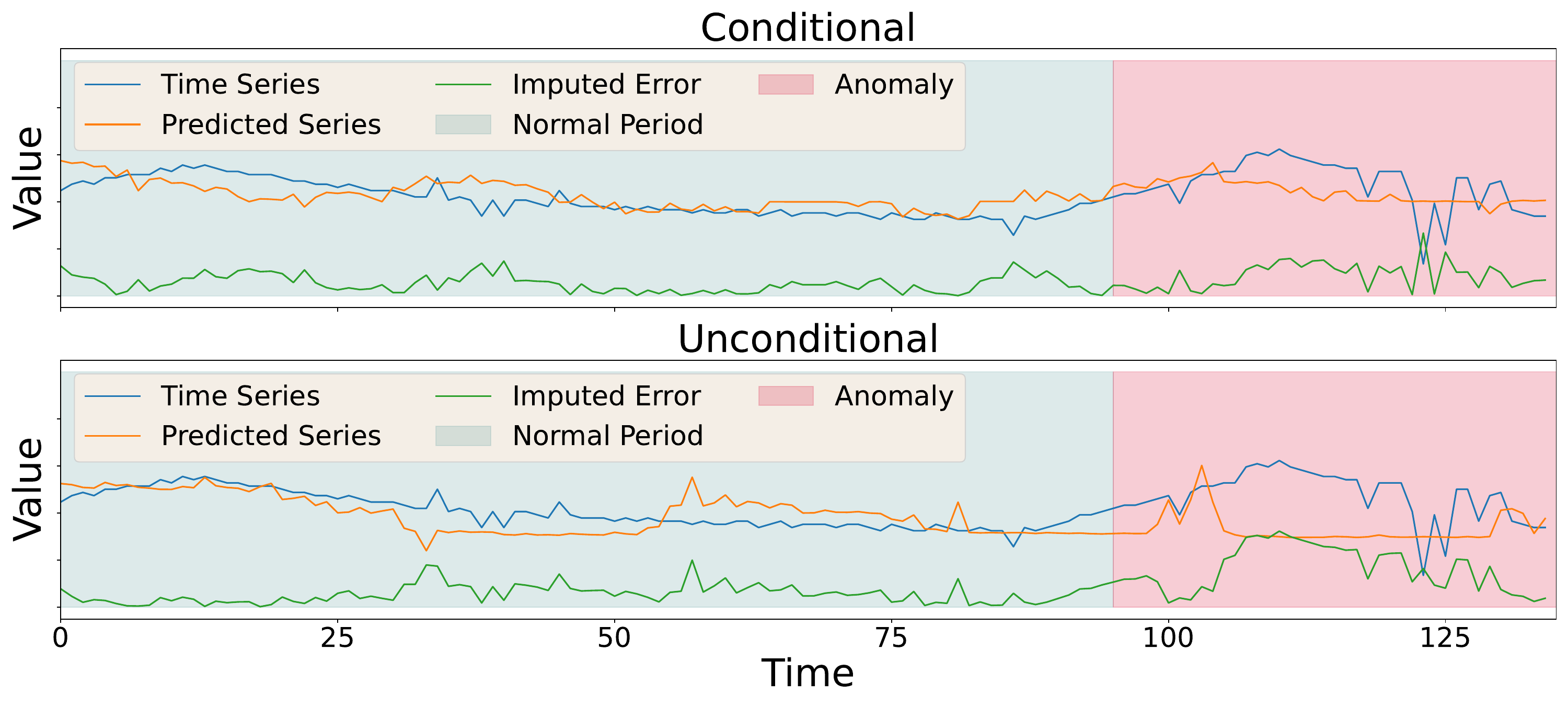}
\vspace*{-1.5em}
\caption{Example cases of conditional/unconditional diffusion models for time series anomaly detection.
\label{fig:conditional_case}}
\vspace*{-1.5em}
\end{figure}

To address this issue, we employ unconditional imputed diffusion models, which utilize the forward noise $\epsilon^{\mathcal{M}_1}_{1:T}$ as a reference for unmasked data input, rather than directly feeding the data values. By using the forward noise, we avoid explicitly revealing the exact values, even when anomaly points are unmasked. However, the forward noise still provides indirect information about the unmasked data, serving as a weak hint for the model. The unmasked data can be perfectly recovered in the reverse process by subtracting the noise from the observed values step-by-step. The lower subplot in Fig.~\ref{fig:conditional_case} demonstrates the application of an unconditional diffusion model. A notable distinction from the conditional model (upper subplot) is the substantial difference in imputed error between normal and abnormal data points. This significant gap in imputed error values provides a distinct boundary for the thresholding approach, which improves the anomaly detection performance.

We denote the noise added to the unmasked input from step $t-1$ to $t$ as $\epsilon^{\mathcal{M}_1}_{t}$. Note that $\epsilon^{\mathcal{M}_1}_{t}$ is drawn from the same Gaussian distribution in Eq.~(\ref{eq:forward}), and serves as the reference for unmasked data in the reverse inference process. Similar to Eq.~(\ref{eq:reverse}), the unconditional imputed diffusion models estimate the masked values in a reverse denoising fashion but condition on the $\epsilon^{\mathcal{M}_1}_{t}$ as additional input. 
Traditional diffusion models lack the capability to incorporate conditional information $\epsilon^{\mathcal{M}_1}_{t}$ during the denoising process. Consequently, an enhancement is required in order to extend the estimation in \rv{Eq.~(\ref{eq:ddpm})} to accommodate conditional information. This can be achieved by modifying the estimation as follows:
\begin{align}
\mathbf{\mu}_\Theta\left (\mathcal{X}^{\mathcal{M}_0}_{t}, t \mid \epsilon^{\mathcal{M}_1}_{t}\right ) &= \mathbf{\mu}\left (\mathcal{X}^{\mathcal{M}_0}_{t}, t, \epsilon_\Theta\left (\mathcal{X}^{\mathcal{M}_0}_{t}, t \mid \epsilon^{\mathcal{M}_1}_{t}\right )\right ), \label{eq:mu} \\ 
\mathbf{\Sigma}_\Theta\left (\mathcal{X}^{\mathcal{M}_0}_{t}, t \mid \epsilon^{\mathcal{M}_1}_{t}\right ) &= \Sigma\left (\mathcal{X}^{\mathcal{M}_0}_{t}, t \right ).\label{eq:sigma}
\end{align}
By utilizing the denoising function $\epsilon_\Theta$ and the forward noise for unmasked value $\epsilon^{\mathcal{M}_1}_{t}$, we can leverage the reverse denoising process of imputed diffusion models to infer the masked values $\mathcal{X}^{\mathcal{M}_0}$. This is accomplished by sampling from the distribution of $\mathcal{X}^{\mathcal{M}_0}_{t}$. In contrast to anomaly detection methods based on reconstruction and forecasting, the integration of additional information offers valuable signals that assist the diffusion model in generating more reliable predictions. This leads to a reduction in output randomness and variance, while maintaining the confidentiality of abnormal data values. Consequently, it enhances the performance and robustness of subsequent anomaly detection. 

\vspace*{-0.5em}
\subsection{Design of Data Masking\label{sec:masking}}
\begin{figure}[t]
\centering
\includegraphics[width=0.95\columnwidth]{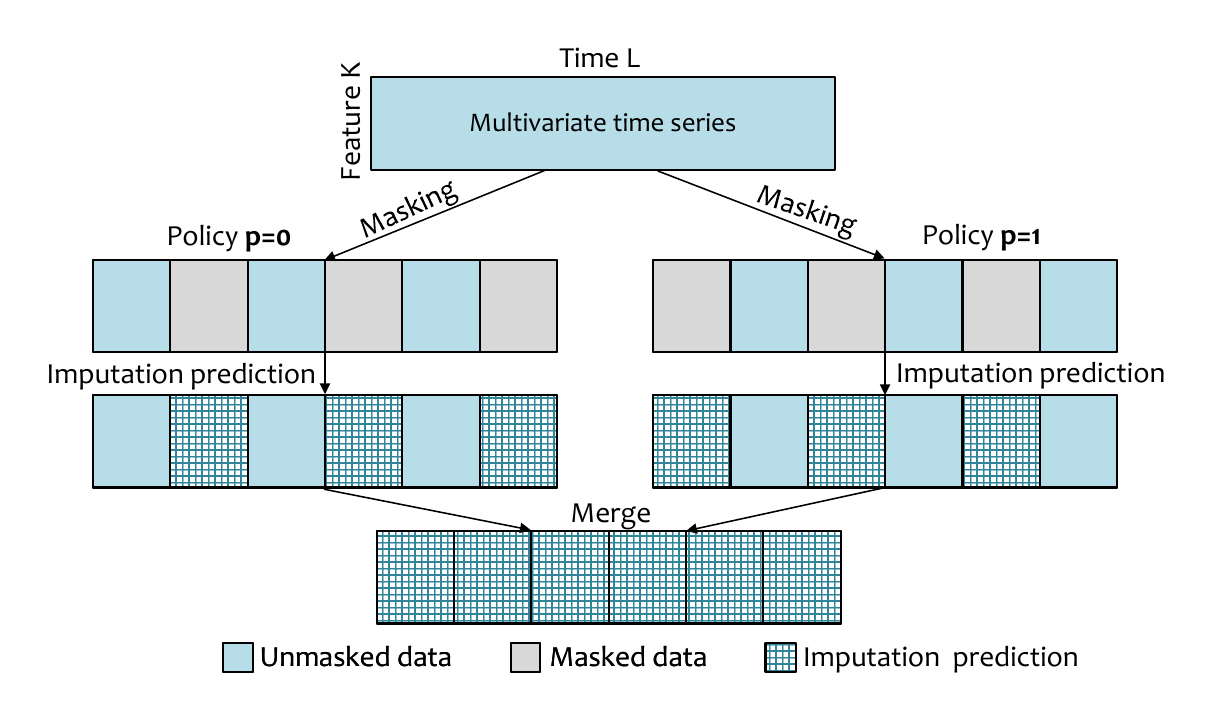}
\vspace*{-1.5em}
\caption{An illustration of the the grating masking and the imputation process under this strategy.
\label{fig:mask}}
\vspace*{-1.5em}
\end{figure}
The \name approach leverages deliberate masking, using a mask $\mathcal{M}$ applied to the time series data, to create unobserved points that require imputation. The choice of the masking strategy plays a crucial role in determining the performance of anomaly detection. In this paper, we compare two masking strategies:
\begin{itemize}[leftmargin=*]
    \item \textbf{Random strategy}: This strategy randomly masks data values in the raw time series with a 50\% probability \cite{tashiro2021csdi}. It provides a straightforward and simple masking technique.
    \item \textbf{Grating strategy}: The grating strategy masks the data at equal intervals along the time dimension, as illustrated in Fig.~\ref{fig:mask}. The raw time series is divided into several windows, with masked and unmasked windows appearing in a staggered manner.
\end{itemize}
For the grating strategy depicted in Fig.~\ref{fig:mask}, two different mask policies indexed by \rv{$p \in \{0, 1\}$} are applied to the same time series, resulting in two imputation instances. These two masks are mutually complementary, ensuring that the masked values in mask $p=0$ are unmasked in mask $p=1$, and vice versa. This guarantees that all data points are imputed by the \name approach, enabling the generation of prediction error signals for anomaly detection. After performing imputation on each masked series individually, the imputation results are merged through simple concatenation. During training and inference, the masking index $p$ is provided to the model, indicating the masking policy applied to reduce ambiguity. This leads to an additional conditional term $p$ on Eq.~(\ref{eq:reverse}) and (\ref{eq:mu}), while the estimation of $\Sigma_\Theta$ in Eq.~(\ref{eq:sigma}) remains unchanged, \ie
\begin{multline}\label{eq:imputed_reverse_m}
    p_{\Theta} \left (\mathcal{X}^{\mathcal{M}_0}_{t-1}\mid\mathcal{X}^{\mathcal{M}_0}_{t}, \epsilon^{\mathcal{M}_1}_{t}\right ) := \mathcal{N}\left (\mathcal{X}^{\mathcal{M}_0}_{t-1}; \mathbf{\mu}_\Theta\left (\mathcal{X}^{\mathcal{M}_0}_{t}, t\mid\epsilon^{\mathcal{M}_1}_{t}, p\right )\right.,\\ \left. \mathbf{\Sigma}_\Theta\left (\mathcal{X}^{\mathcal{M}_0}_{t}, t\mid\epsilon^{\mathcal{M}_1}_{t}, p\right )\mathbf{I}\right ),
\end{multline}
\vspace*{-1.5em}
\begin{align}
\mathbf{\mu}_\Theta\left (\mathcal{X}^{\mathcal{M}_0}_{t}, t \mid \epsilon^{\mathcal{M}_1}_{t}\right ) &= \mathbf{\mu}\left (\mathcal{X}^{\mathcal{M}_0}_{t}, t, \epsilon_\Theta\left (\mathcal{X}^{\mathcal{M}_0}_{t}, t \mid \epsilon^{\mathcal{M}_1}_{t}, p\right )\right ).
\end{align}

The grating strategy introduces a unique characteristic to the imputation, as it can be considered a ``partially'' reconstruction task. This approach offers several advantages: Firstly, it provides additional information that aids in modeling the time series more effectively. By incorporating the partially reconstructed data, the model gains a better understanding of the underlying patterns and correlations. Secondly, the utilization of the grating strategy allows for a partial glimpse into the future values of the time series within the masked window, akin to forecasting techniques. This enables to improve timeliness in detecting anomalies, as it provides insights into the potential future trajectory of the time series data. 

\vspace*{-0.5em}
\subsection{Training Process of \name\label{sec:training}}

\begin{figure}[t]
\centering
\vspace*{-1em}
\includegraphics[width=0.9\columnwidth]{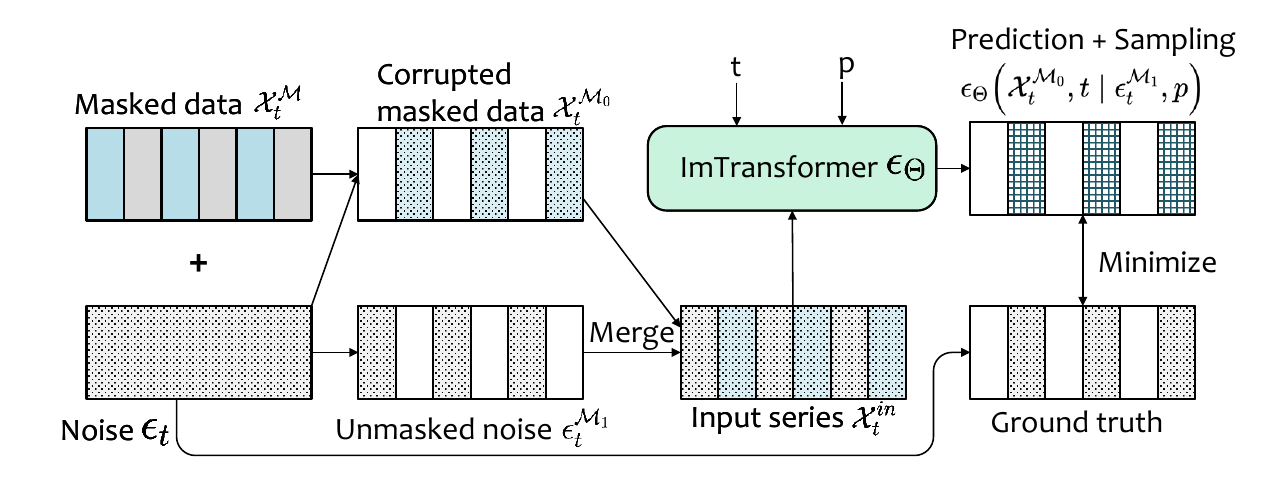}
\vspace*{-2em}
\caption{The training process of \name.
\label{fig:train}}
\vspace*{-1em}
\end{figure}

\begin{figure}[t]
\centering
\includegraphics[width=1\columnwidth]{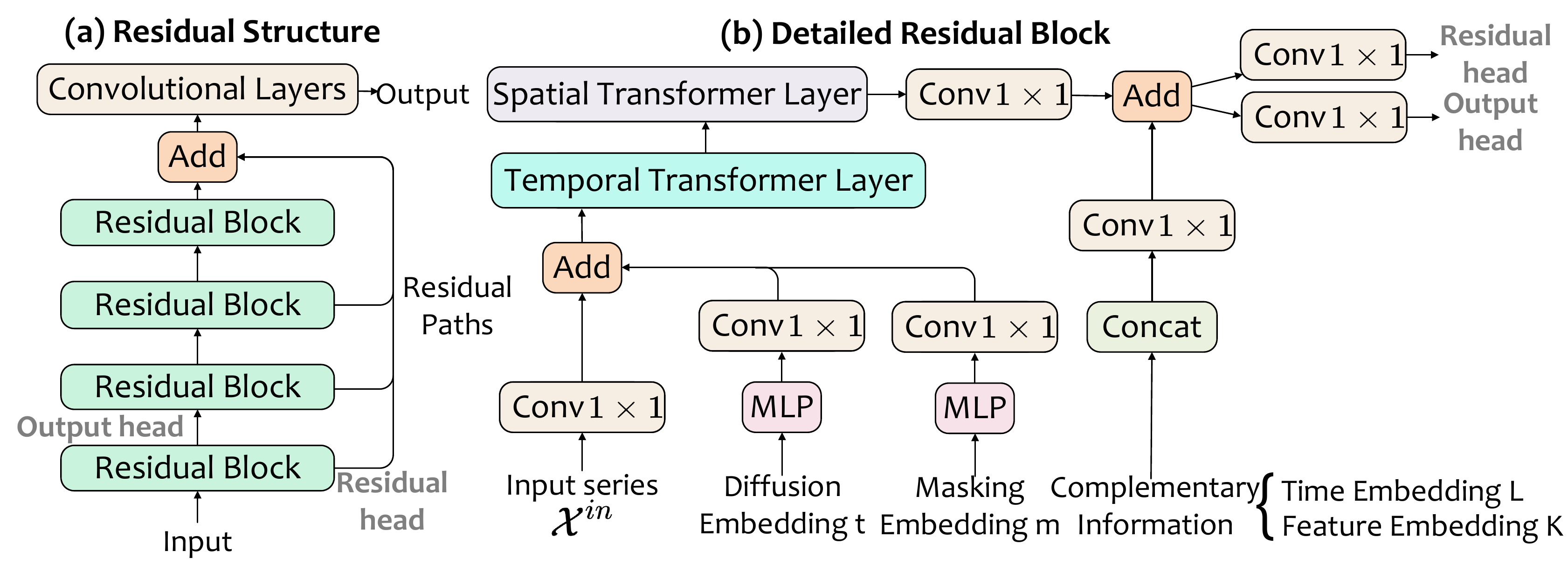}
\vspace*{-2.5em}
\caption{The \model architecture, with (a) the residual structure; and (b) the details of a residual block. 
\label{fig:model}}
\vspace*{-1.5em}
\end{figure}

The training process of \name is illustrated in Fig.~\ref{fig:train}. Starting with a given data sample time series $\mathcal{X}$, we generate two masked samples $\mathcal{X}^\mathcal{M}$ using the grating masking strategy discussed in Sec.~\ref{sec:masking}. These masked samples are gradually corrupted by introducing Gaussian noise $\epsilon_t$, resulting in $\mathcal{X}^\mathcal{M}_t$. Our objective is to train a model $\mu_\Theta$ that can effectively denoise $\mathcal{X}^\mathcal{M}_t$ and impute the masked values in a step-by-step manner. Given that \name employs an unconditional diffusion model, the input series $\mathcal{X}^{in}_t$ is divided into two halves, \ie $\mathcal{X}^{in} = \{\mathcal{X}_t^{\mathcal{M}_0}, \epsilon _t^{\mathcal{M}_1}\}$. One half contains the corrupted data within the masked regions, denoted as $\mathcal{X}^{\mathcal{M}_0}_t$, while the other half represents the ground truth forward noise applied to the unmasked regions, denoted as $\epsilon^{\mathcal{M}_1}_t$, serving as a reference. These two data sources are concatenated to form the input $\mathcal{X}^{in}_t$, which is then fed into the dedicated transformer-based model \model for denoising inference, \ie $\epsilon_\Theta\left ( \mathcal{X}^{\mathcal{M}_0}_t, t \mid \epsilon^{\mathcal{M}_1}_t, p \right )$.

The unconditional imputed diffusion models utilize the same parameterization as Eq.~(\ref{eq:ddpm}), with the only difference lying in the form of $\mu_\Theta$, which takes additional inputs of unmasked forward noise $\epsilon^{\mathcal{M}_1}_{t}$ and mask index $p$. We follow the standard training process of DDPM, beginning with sampling Gaussian noise as masked data at step $T$, \ie $\mathcal{X}_{T} = \sqrt{\alpha_T}\mathcal{X}_{0} + (1-\alpha_T)\epsilon$, and optimizing $\epsilon_\Theta$ by minimizing the following loss function:
\begin{align}
    \min_\Theta \mathcal{L}(\Theta) := \min_\Theta \mathbb{E}_{\mathcal{X}_0\sim q(\mathcal{X}_0), \epsilon\sim\mathcal{N}(\mathbf{0}, \mathbf{I}),t}||\epsilon -\epsilon _\Theta(\mathcal{X}^{\mathcal{M}_0}_{t},t \mid \epsilon^{\mathcal{M}_1}_{t}, p)||^2.
\end{align}
Once trained, we can utilize the diffusion model to infer the masked values given a random Gaussian noise $\mathcal{X}_T^{\mathcal{M}_0}$, as well as the forward noise sequence added to the unmasked data $\epsilon^{\mathcal{M}_1}_{1:T}$.

\vspace*{-0.5em}
\subsection{Imputation with \model \label{sec:label}}
Drawing inspiration from the studies conducted in \cite{han2021transformer, kongdiffwave, tashiro2021csdi}, which employ hierarchical structures of transformers \cite{vaswani2017attention} to capture temporal correlations and interactions among variables, we introduce \model, a specialized architecture designed for MTS imputation, as illustrated in Fig.~\ref{fig:model}. It comprises a series of stacked residual blocks, with each  containing dedicated components that process the feature and temporal dimensions separately.

The \model model incorporates four distinct groups of input data: \emph{(i)} the input time series $\mathcal{X}_t^{in}$, \emph{(ii)} diffusion embedding that encodes information related to the current diffusion step $t$, \emph{(iii)} masking embedding that encodes the masking group $p$ of the current data, and \emph{(iv)} complementary information that embeds the dimensional information of time $l$ and feature $k$. Each of these groups of data is individually processed by convolutional and/or multilayer perceptron layers to ensure a consistent dimensionality. The embeddings of the inputs are then combined into a single tensor and further processed by a temporal and a spatial transformer layer.

The temporal transformer plays a crucial role in capturing the temporal dependencies within the time series \cite{zhou2021informer}. It enables the dynamic weighting of feature values at different time steps and takes into account the masked status of features. The attention mechanism employed in the temporal transformer provides the necessary flexibility for this purpose. Additionally, a 1-layer spatial transformer is employed to capture the interdependencies between different variables at each time step. This spatial transformer allows for adaptive weighting and facilitates interaction between variables. The output of the spatial transformer is combined with the complementary information, creating a residual head for skip connection, as illustrated in Fig.~\ref{fig:model}. Both the spatial and temporal transformers play crucial roles in the imputation and anomaly detection tasks, as the feature and temporal dimensions may contribute differently to the predictions \cite{tuli2022tranad}, which can be learned by the attention mechanism \cite{vaswani2017attention}. The incorporation of a residual structure \cite{he2016deep} further enhances the model capacity by facilitating gradient propagation.

\vspace*{-0.5em}
\subsection{Ensemble Anomaly Inference}

\begin{algorithm}[tb]
  \caption{The ensemble inference process of \name.\label{alg:ensemble}}
  \begin{algorithmic}[1]
    \Inputs{Masked data input series $\mathcal{X}_t^{in}$, masking tensors $\mathcal{M}$, a trained denoising model $\epsilon_\Theta$, the forward ground truth noise on unmasked region $\epsilon^{\mathcal{M}_1}_{1:T}$, total denosing step $T$.}
    \Initialize{Initial noise vector $\mathcal{X}_T$.}\label{ln:initial}
    \For{$t = T$ to $1$} \label{ln:loopd}
        \State{Construct two input series $\mathcal{X}_t^{in} = \{\mathcal{X}_t^{\mathcal{M}_0}, \mathcal{\epsilon}_t^{\mathcal{M}_1}\}$ with} \Statex{\hspace*{1.5em}masking $\mathcal{M}$. }\label{ln:input}
           \State{Predicting $\mu_\Theta$, $\Sigma_\Theta$ using the denosing model $\epsilon_\Theta$.}\label{ln:predict}
           \State{Sampling using equation~(\ref{eq:imputed_reverse_m}) and obtain predicted $\mathcal{X}_{t-1}$.}\label{ln:sample}
           \State{Compute prediction error $E_t = || \mathcal{X} - \mathcal{X}_{t-1} ||^2$.}\label{ln:error}
	\EndFor  
    \For{$t = T$ to $1$} \label{ln:loopd2}
        \State{Computing the anomaly prediction label $Y_t$ using Eq.~(\ref{eq:steplabel}).}
    \EndFor
    \State{Aggregating the voted anomaly prediction $\mathcal{V}_l = \sum_{t=1}^T y_{t, l}$.}
    \State{Computing the final anomaly prediction $y_l = \mathbbm{1}(\mathcal{V}_l > \xi)$.}
    \end{algorithmic}
\end{algorithm}
\setlength{\textfloatsep}{2pt}

Traditional anomaly detection models typically rely on a single signal, \ie the prediction error, to determine the anomaly label for testing data. However, relying solely on one signal can lead to unrobust predictions, as the prediction error can be subjective to stochasticity and affected by various random factors. \rv{The presence of anomalous data within the training set further raises concerns about the robustness requirement.} To address this limitation, we leverage the unique advantage of diffusion models. Unlike traditional models that provide a single-shot prediction, imputed diffusion models progressively denoise the masked data over $T$ steps. This results in at least $T$ intermediate outputs, each having the same dimension as the original time series, which is not available in traditional models. Although these intermediate outputs are not fully denoised, they converge towards the same imputation objective and offer different perspectives on the time series modeling. By appropriately utilizing these outputs, we can uncover the step-by-step reasoning of \name and utilize them as additional signals to enhance the robustness and accuracy of anomaly detection.

\begin{figure}[t]
\centering
\includegraphics[width=\columnwidth]{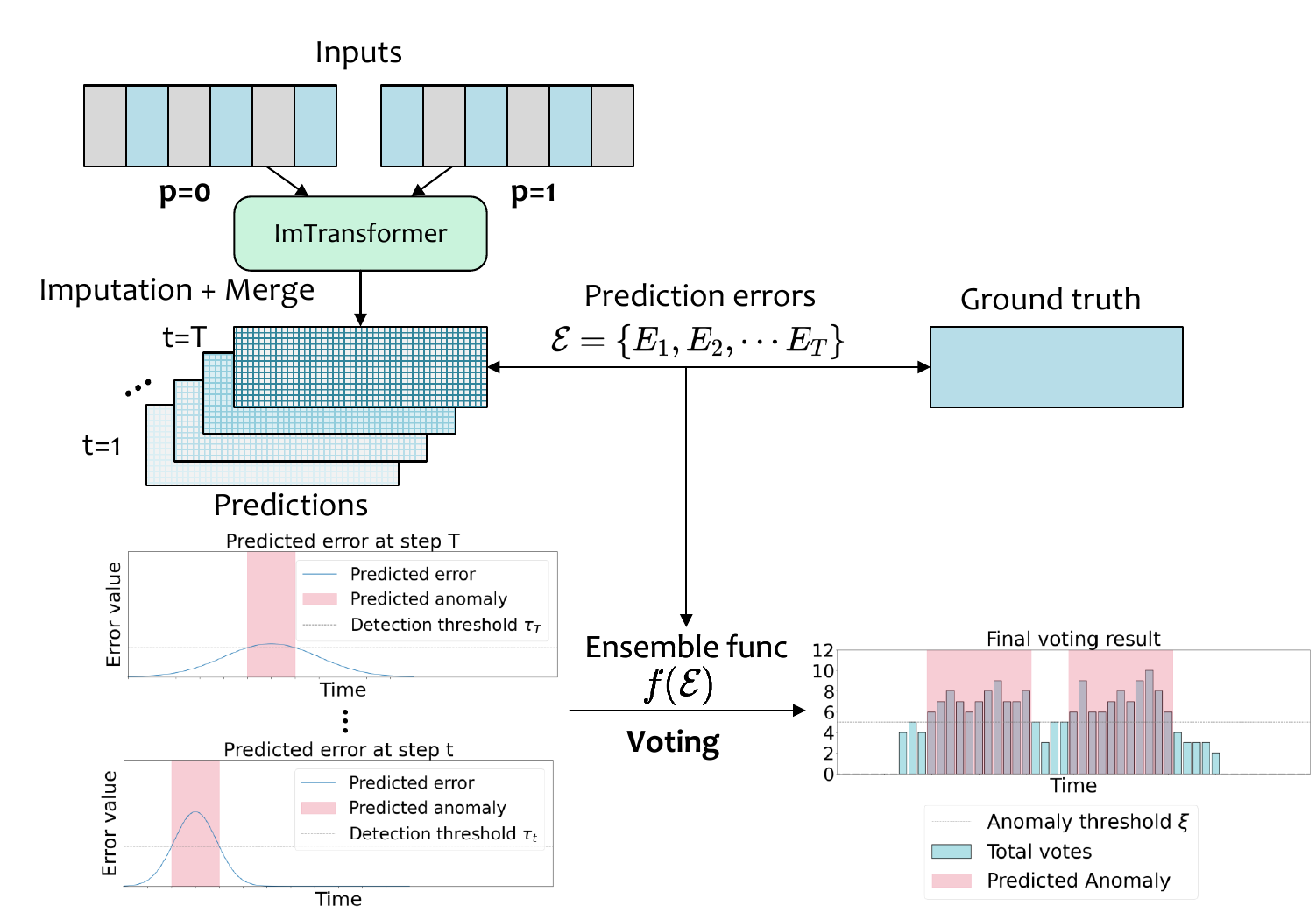}
\vspace*{-2.5em}
\caption{The ensemble anomaly inference  of \name.
\label{fig:infer}}
\vspace*{-1em}
\end{figure}

\name utilizes the prediction error at each denoising step $t$, denoted as $\mathcal{E} = \{E_1, E_2, \cdots, E_T\}$, as input and ensembles them using a function $f(\mathcal{E})$ to determine the final anomaly labels. $E_t$ denotes the prediction error tensor for the imputed output at denoising step $t$, and it has the same dimension as the original time series $\mathcal{X}$. The ensemble anomaly inference algorithm is presented in Algorithm~\ref{alg:ensemble}, and Fig.~\ref{fig:infer} provides an illustration of the process. At each denoising step, \name generates a prediction using the denoising model $\epsilon_\Theta$ and computes the prediction error $E_t$ with respect to the ground truth time series $\mathcal{X}$. The set $\mathcal{E}$ collects the prediction errors at each denoising step, and an ensemble function $f(\mathcal{E})$ is employed to leverage the all-step errors to obtain the final voting signal $\mathcal{V}$ for determining the anomaly label, \ie $\mathcal{V} = f(\mathcal{E})$. 

\noindent \textbf{The design of the ensemble function.}  \name utilizes a voting ensemble mechanism \cite{dietterich2000ensemble} to strengthen the overall anomaly detection process by aggregating anomaly predictions from each denoising step. At each denoising step $t$, the anomaly prediction label $Y_t$ is determined using the following equation:
\begin{align}\label{eq:steplabel}
    Y_t = \mathbbm{1}(E_t \geq \tau_t), \ \mathrm{where}\ \tau_t = \frac{\sum E_T }{\sum E_t} \cdot \tau_T.
\end{align}
Here, $\tau_T$ represents the upper percentile of imputed errors at the final denoising step $T$. The rationale behind this design is to utilize the imputed error at the last step as a baseline and use it as an indicator of imputation quality. The rescaling ratio $\frac{\sum E_T }{\sum E_t}$ measures the imputation quality at each step $t$. If the ratio is small, it indicates poor imputation quality, and therefore the upper percentile of imputed errors for determining the anomaly label is reduced. In this case, only the label for the timestamp with the highest imputed error and high confidence is retained. Conversely, if the ratio is small, it suggests good imputation quality, and the error threshold for anomaly detection is relaxed. This dynamic adjustment of the threshold allows for adaptability based on the quality of imputation.

Using Eq.~(\ref{eq:steplabel}), we derive the step-wise anomaly predictions $Y_t = \{y_{t, 1}, \cdots, y_{t, L}\}$, where $y_{t, l}=1$ indicates that the data at time step $l$ is predicted as an anomaly using the imputation at diffusion step $t$, and $y_{t, l}=0$ otherwise. To determine the final anomaly prediction at each time step $l$, we employ a voting mechanism. The voting signal $\mathcal{V}_l$ represents the total number of anomaly votes received at time step $l$, given by $\mathcal{V}_l = \sum_{t=1}^T y_{t, l}$. If a time step receives more than $\xi$ votes as an anomaly across all denoising steps, it is marked as a final anomaly, denoted as $y_l = \mathbbm{1}(\mathcal{V}_l > \xi)$. To optimize inference efficiency and ensure correctness, we sample every 3 steps from the last 30 denoising steps for the voting process. This voting mechanism strengthens the \name framework by utilizing the intermediate imputed outputs as additional signals. \emph{This is unique to diffusion models, as they generate predictions progressively}.





\vspace*{-0.5em}
\section{Offline Evaluation}
\begin{table}[t]
\caption{Hyperprameter setting of \name. \label{tab:hyperprameter}}
\vspace*{-1.em}
\begin{tabular}{ll}
\hline
Hyperprameter                                & Value                                                                              \\ \hline
Number of grating masked windows                        & 5                                                                                  \\
Number of grating unmasked windows                      & 5                                                                                  \\
Detection window size                      & 100                                                                                  \\
\model residual block         & 4                                                                                  \\
\model hidden layer dimension & 128                                                                                \\
Denosing step $T$                            & 50                                                                                 \\ \hline
\end{tabular}
\end{table}
We conducted a comprehensive offline evaluation of the \name for MTS anomaly detection. The evaluation aimed to address the following research questions (RQs):
\begin{itemize}[leftmargin=*]
    \item \textbf{RQ1:} How does \name perform compare to state-of-the-art methods in MTS anomaly detection?
    \item \textbf{RQ2:} How effective are each specific design in \name?
    \item \textbf{RQ3:} What insights can be gained from each mechanism employed in \name?
\end{itemize}

\noindent \textbf{Implementation.} The \name framework is implemented using the \texttt{PyTorch} framework \cite{paszke2019pytorch} and trained on a GPU cluster comprising multiple NVIDIA RTX 1080ti, 2080ti, and 3090 accelerators.
Table~\ref{tab:hyperprameter} presents the key hyperparameters used in \name. The detection thresholds $\tau$ for the MSL dataset vary across different subsets, while a fixed value of 0.02 is employed for the other datasets. The voting threshold $\xi$ is dataset-dependent and is specified in the provided code link. As for the baseline models, their hyperparameters and detection thresholds are set based on the information provided in their respective original papers. In cases where these details were not explicitly mentioned, a grid search was conducted to determine the optimal values.

\vspace*{-0.5em}
\subsection{Datasets, Baselines \& Evaluation Metrics}
We test the performance of \name using 6 publicly available MTS anomaly detection datasets, namely SMD \cite{su2019robust}, PSM \cite{abdulaal2021practical}, MSL \cite{hundman2018detecting}, SMAP \cite{hundman2018detecting}, SWaT \cite{mathur2016swat} and GCP \cite{ma2021jump}. In order to ensure the completeness of the experiment, we trained and evaluated the \name on all subsets of the aforementioned dataset, rather than selectively choosing non-trivial sequences as done in \cite{tuli2022tranad}. This may lead to differences in the evaluation metrics compared to the results reported in the original paper.

We evaluate the performance of \name by comparing it with 10 state-of-the-art MTS anomaly detection models:
\emph{(i)} Isolation forest (IForest) \cite{liu2012isolation} separates the anomaly data point with others for detection.
\emph{(ii)} BeatGAN \cite{zhou2019beatgan} utilizes generative adversarial networks (GANs) \cite{goodfellow2020generative, zhang2017zipnet} to reconstruct time series and detect anomalies.
\emph{(iii)} LSTM-AD \cite{malhotra2015long} employs LSTM \cite{hochreiter1997long, zhang2021cloudlstm} to forecast future values and uses the prediction error as an indicator of anomalies.
\emph{(iv)} InterFusion \cite{li2021multivariate} captures the interaction between temporal information and features to effectively identify inter-metric anomalies.
\emph{(v)} OmniAnomaly \cite{su2019robust} combines GRU \cite{chung2014empirical} and VAE \cite{park2018multimodal} to learn robust representations of time series and utilizes the Peaks-Over-Threshold (POT) \cite{siffer2017anomaly} method for threshold selection.
\emph{(vi)} GDN \cite{deng2021graph} introduces graph neural networks into anomaly detection and leverages meta-learning methods to combine old and new knowledge for anomaly identification.
\emph{(vii)} MAD-GAN \cite{li2019mad} employs GANs \cite{goodfellow2020generative, zhang2017zipnet} to recognize anomalies by reconstructing testing samples from the latent space.
\emph{(viii)} MTAD-GAT \cite{zhao2020multivariate} utilizes Graph Attention Network (GAT) \cite{velivckovicgraph} to model MTS and incorporates forecasting-based and reconstruction-based models to improve representation learning \cite{zhang2020microscope}.
\emph{(ix)} MSCRED \cite{zhang2019deep} uses ConvLSTM networks \cite{shi2015convolutional, zhang2018long, zhang2019multi} to capture correlations among MTS and operates as an anomaly detector.
\emph{(x)} TranAD \cite{tuli2022tranad} leverages transformer models to perform anomaly inference by considering the broader temporal trends in the data.
We benchmark \name against these models to evaluate its performance in MTS anomaly detection.


In line with previous studies \cite{tuli2022tranad, li2021multivariate, zhao2020multivariate}, we evaluate the anomaly detection accuracy of both the baseline models and our proposed \name using precision, recall, and F1 score. Note that we conducted 6 independent runs for each baseline model and \name, and report the average performance. Additionally, we provide the standard deviation of the F1 score (F1-std) in the 6 runs to assess the stability and robustness of all methods examined in this investigation. 
We also utilize the R-AUC-ROC evaluation metric introduced in \cite{paparrizos2022volume} to provide a threshold-independent accuracy assessment tailored to range-based anomalies. This metric mitigates the bias introduced by threshold selections and offers a different perspective on the performance of anomaly detection methods by using continuous buffer regions.
Further, we utilize the Average Sequence Detection Delay (ADD) metric proposed in \cite{doshi2022reward} to evaluate the speed and timeliness of anomaly detection provided by each approach. The ADD metric is defined as follows:
\begin{align}\label{eq:add}
\mathrm{ADD} = \frac{1}{S} \sum_{i=1}^S (\mathcal{T}_i - \varrho_i),
\end{align}
where $\varrho_i$ represents the start time of anomalous event $i$, $\mathcal{T}_i \geq \varrho_i$ denotes the corresponding detection delay time by the anomaly detector, and $S$ indicates the total number of anomalous events. A smaller value of ADD indicates a more timely detection of anomalies, which is crucial in real-world detection scenarios.

\begin{table*}[t]
\centering
\caption{The Precision (P), Recall (R), F1 and R-AUC-ROC of all anomaly detectors on benchmark datasets. The average values of P, R, F1 and R-AUC-ROC were calculated from 6 individual runs, while F1-std. is the standard deviation across the 6 runs. \label{tab:performance}}
\vspace*{-1.em}
\resizebox{1.\textwidth}{!}{
\begin{tabular}{lccccccccccccccc}
\hline
\multirow{2}{*}{Method} & \multicolumn{5}{c}{SMD}                                                                 & \multicolumn{5}{c}{PSM}                                                                 & \multicolumn{5}{c}{SWaT}                                                                           \\ \cline{2-16} 
                        & P               & R               & F1              & F1-std.         & \footnotesize{R-AUC-PR}    & P               & R               & F1              & F1-std.         & \footnotesize{R-AUC-PR}    & P               & R               & F1              & F1-std.                    & \footnotesize{R-AUC-PR}    \\ \hline
IForest         & 0.2030          & 0.2130          & 0.1799          & 0.0138          & 0.0257          & 0.6630          & 0.4919          & 0.5641          & 0.0070          & 0.2058          & 0.9764          & 0.6650          & 0.7907          & 0.0020                     & 0.0685          \\
BeatGAN                 & 0.9013          & 0.8894          & 0.8797          & 0.0058          & 0.3200          & 0.9204          & 0.8767          & 0.8975          & 0.0178          & 0.3453          & 0.9606          & 0.7020          & 0.8107          & 0.0022                     & 0.3215          \\
LSTM-AD                 & 0.3361          & 0.3229          & 0.2639          & 0.0123          & 0.0399          & 0.9050          & 0.7707          & 0.8313          & 0.0036          & 0.2561          & \textbf{0.9925} & 0.6737          & 0.8026          & 0.0013                     & 0.3118          \\
InterFusion             & 0.8815          & 0.9071          & 0.8772          & 0.0226          & 0.3012          & 0.9533          & 0.9128          & 0.9326          & 0.0036          & 0.1896          & 0.8683          & 0.8530          & 0.8600          & 0.0309                     & 0.1477          \\
OmniAnomaly             & 0.8751          & 0.9052          & 0.8775          & 0.0083          & 0.2525          & 0.9551          & 0.8859          & 0.9191          & 0.0060          & 0.3718          & 0.9749          & 0.7500          & 0.8470          & 0.0271                     & \textbf{0.3722} \\
GDN                     & 0.8460          & 0.7862          & 0.7865          & 0.0109          & 0.1637          & 0.8750          & 0.8385          & 0.8564          & \textbf{0.0000} & 0.3230          & 0.1311          & 0.0585          & 0.0808          & \textbf{0.0009}            & 0.1318          \\
MAD-GAN                 & 0.8851          & 0.9045          & 0.8803          & 0.0384          & 0.2295          & 0.8596          & 0.8838          & 0.8698          & 0.0339          & 0.4416          & 0.7918          & 0.5423          & 0.6385          & 0.3048                     & 0.2633          \\
MTAD-GAT                & 0.8836          & 0.8330          & 0.8463          & 0.0316          & 0.3006          & 0.8763          & 0.8725          & 0.8744          & 0.0000          & 0.4116          & 0.8468          & 0.8224          & 0.8344          & 0.0067                     & 0.3196          \\
MSCRED                  & 0.8567          & 0.9038          & 0.8426          & \textbf{0.0002} & 0.2601          & 0.9555          & 0.6857          & 0.7965          & 0.0102          & 0.3846          & 0.4823          & 0.4065          & 0.4407          & 0.3408                     & 0.1668          \\
TranAD                  & 0.8906          & 0.8982          & 0.8785          & 0.0023          & 0.2941          & 0.9506          & 0.8951          & 0.9220          & 0.0045          & 0.3994          & 0.7025          & 0.7266          & 0.6886          & 0.1089                     & 0.1670          \\
\name    & \textbf{0.9520} & \textbf{0.9509} & \textbf{0.9488} & 0.0039          & \textbf{0.3821} & \textbf{0.9811} & \textbf{0.9753} & \textbf{0.9781} & 0.0072          & \textbf{0.4711} & 0.8988          & \textbf{0.8465} & \textbf{0.8709} & 0.0124                     & 0.1939          \\ \hline\hline
\multirow{2}{*}{Method} & \multicolumn{5}{c}{SMAP}                                                                & \multicolumn{5}{c}{MSL}                                                                 & \multicolumn{5}{c}{GCP}                                                                            \\ \cline{2-16} 
                        & P               & R               & F1              & F1-std.         & \footnotesize{R-AUC-PR}    & P               & R               & F1              & F1-std.         & \footnotesize{R-AUC-PR}    & P               & R               & F1              & F1-std.                    & \footnotesize{R-AUC-PR}    \\ \hline
IForest         & 0.2886          & 0.7671          & 0.4163          & 0.0026          & 0.1096          & 0.6059          & 0.5328          & 0.5334          & 0.0309          & 0.0942          & 0.8055          & 0.7385          & 0.7370          & 0.0120                     & 0.1558          \\
BeatGAN                 & 0.8915          & 0.6781          & 0.7663          & 0.0162          & \textbf{0.1303} & 0.7782          & 0.8512          & 0.8102          & 0.0342          & 0.1421          & 0.9865          & 0.9630          & 0.9717          & 0.0074                     & 0.2414          \\
LSTM-AD                 & 0.7841          & 0.5630          & 0.6533          & 0.0382          & 0.1099          & 0.7330          & 0.5745          & 0.6378          & 0.1473          & 0.1066          & 0.9591          & 0.9575          & 0.9553          & 0.0013                     & 0.2610          \\
InterFusion             & 0.8788          & 0.7704          & 0.8204          & 0.0077          & 0.1457          & 0.7688          & \textbf{0.9464} & 0.8442          & 0.0330          & 0.1083          & 0.9361          & 0.9720          & 0.9092          & 0.0005                     & 0.2846          \\
OmniAnomaly             & 0.8407          & \textbf{0.9674} & 0.8995          & 0.0078          & 0.0978          & 0.8321          & 0.8125          & 0.8221          & 0.0121          & 0.1290          & 0.9572          & 0.9796          & 0.9668          & 0.0027                     & 0.2029          \\
GDN                     & 0.9689          & 0.5401          & 0.6936          & 0.0037          & 0.0961          & 0.8668          & 0.8072          & 0.8360          & 0.0004          & 0.1295          & 0.9648          & 0.9628          & 0.9589          & 0.0011                     & 0.2096          \\
MAD-GAN                 & 0.9547          & 0.5474          & 0.6952          & 0.0013          & 0.0990          & 0.7047          & 0.7841          & 0.7423          & \textbf{0.0000} & 0.1301          & 0.9766          & 0.9558          & 0.9605          & 0.0055                     & 0.1867          \\
MTAD-GAT                & \textbf{0.9718} & 0.5259          & 0.6824          & \textbf{0.0012} & 0.1083          & 0.7321          & 0.7616          & 0.7432          & 0.0200          & 0.1278          & 0.9490          & 0.9523          & 0.9461          & \multicolumn{1}{c}{0.0047} & 0.2210          \\
MSCRED                  & 0.4107          & 0.8604          & 0.2712          & 0.0625          & 0.1042          & 0.5008          & 0.6088          & 0.4899          & 0.0788          & 0.1090          & 0.9754          & 0.9735          & 0.9712          & \textbf{0.0006}            & 0.2068          \\
TranAD                  & 0.8224          & 0.8502          & 0.8360          & 0.0090          & 0.1077          & \textbf{0.8951}           & 0.9297          & \textbf{0.9115} & 0.0051          & 0.1057          & 0.9472          & 0.9812          & 0.9631          & 0.0030                     & 0.2026          \\
\name    & 0.8771          & 0.9618          & \textbf{0.9175} & 0.0095          & 0.1105          & 0.8930 & 0.8638          & 0.8779          & 0.0152          & \textbf{0.2381} & \textbf{0.9771} & \textbf{0.9825} & \textbf{0.9774} & 0.0014                     & \textbf{0.3957} \\ \hline
\end{tabular}}
\vspace*{-1.em}
\end{table*}

\begin{table}[t]
\centering
\caption{P, R, F1, F1-std and R-AUC-ROC performance of all anomaly detectors averaged over six benchmark datasets. \label{tab:avg}}
\vspace*{-1.em}
\resizebox{\columnwidth}{!}{
\begin{tabular}{lccccc}
\hline
Method               & P               & R               & F1              & F1-std.         & \footnotesize{R-AUC-PR}    \\ \hline
IForest              & 0.5904          & 0.5680          & 0.5369          & 0.0114          & 0.1099          \\
BeatGAN              & 0.9064          & 0.8267          & 0.8560          & 0.0139          & 0.2501          \\
LSTM-AD              & 0.7850          & 0.6437          & 0.6907          & 0.0340          & 0.1809          \\
InterFusion          & 0.8811          & 0.8936          & 0.8739          & 0.0164          & 0.1962          \\
OmniAnomaly          & 0.9058          & 0.8834          & 0.8887          & 0.0107          & 0.2377          \\
GDN                  & 0.7754          & 0.6656          & 0.7020          & \textbf{0.0028} & 0.1756          \\
MAD-GAN              & 0.8621          & 0.7697          & 0.7978          & 0.0640          & 0.2250          \\
MTAD-GAT             & 0.8766          & 0.7946          & 0.8211          & 0.0107          & 0.2481          \\
MSCRED               & 0.6969          & 0.7398          & 0.6353          & 0.0822          & 0.2053          \\
TranAD               & 0.8681          & 0.8802          & 0.8666          & 0.0221          & 0.2128          \\
\name & \textbf{0.9298} & \textbf{0.9301} & \textbf{0.9284} & 0.0083       & \textbf{0.2986} \\ \hline
\end{tabular}
}
\end{table}


\vspace*{-0.5em}
\subsection{Anomaly Detection Performance (RQ1)}
\subsubsection{Accuracy Performance} 
We first present the precision, recall, F1 and R-AUC-PR performance of \name and the baseline methods in Table~\ref{tab:performance} for each of the six datasets considered in this study. Please note that all the results presented in the table are the average values obtained from 6 individual runs, which allows us to assess the robustness of each detector. Additionally, the F1-std. (standard deviation) provides an indication of the variability of the F1 scores across these runs. represents the standard deviation across the 6 runs. Additionally, Table~\ref{tab:avg} displays the average performance across all six datasets.  Notably, \name overall demonstrates exceptional performance in terms of all evaluation metrics, namely precision (92.98\%), recall (93.01\%), and F1 score (92.84\%) and R-AUC-PR (29.86\%). It achieves the highest average scores across six datasets, surpassing the performance of the other baseline methods. In particular, \name exhibits at least a 2.4\% increase in precision, a 4.67\% increase in recall, a 3.97\% increase in F1 score and a 4.85\% increase in and R-AUC-PR compared to the other baselines. These results demonstrate the effectiveness of the imputation approach and diffusion models employed in \name.

Furthermore, despite that diffusion models require sampling at every denoising step, introducing randomness, the F1-std (0.0083) calculated from 6 independent runs remains relatively small compared to other baselines, ranking second lowest among all approaches. This indicates the remarkable robustness of \name. It can be attributed to two key design elements in \name: \emph{(i)} the imputation methods leverage neighboring information for self-supervised modeling, reducing prediction uncertainty, and \emph{(ii)} the dedicated ensemble mechanism aggregates votes for step-wise anomaly inference, further reducing prediction variance. We provide a more detailed ablation study in Sec.~\ref{sec:unsupervised_ablation} and~\ref{sec:ensemble_ablation}. 

Upon closer examination of the dataset-specific performance in Table~\ref{tab:performance}, we observe that \name achieves the highest F1 score in 5 out of the 6 datasets. The exception is the MSL dataset, where TranAD outperforms \name. This can be attributed to the fact that it is specifically designed to capture the internal correlations across different dimensions, which are the prominent characteristics of the MSL dataset. \rv{A plausible solution to reinforce \name is to explicitly model these dependencies through hierarchical inter-metric embedding, as employed in InterFusion \cite{li2021multivariate}.} However, \name also takes a different approach by leveraging the exceptional self-supervised learning ability of diffusion models \rv{and the spatial transformer in \model} to capture correlations and provide a more general solution across various datasets. This enables \name to achieve competitive performance in most datasets and surpass other baselines. Furthermore, we observe that \name also achieves the highest R-AUC-PR in 4 out of the 6 datasets. 
This highlights the robustness of \name to threshold selection and its consistent ability to deliver accurate predictions in detecting range anomalies.

\rv{However, in the SWaT and SMAP datasets, we observe a notable reduction in precision for \name compared to several baselines. This can be attributed to a slight overfitting exhibited by \name on these specific datasets, which leads to increased errors in normal data. Consequently, applying a fixed error threshold results in the identification of more false anomalies, thereby compromising precision. A potential solution could involve the implementation of dynamic thresholding approaches \cite{hundman2018detecting} to achieve a better balance between precision and recall. Moreover, mitigating overfitting can be achieved by reducing the complexity of the \model. These considerations are reserved for future work.}


Notably, the performance improvements achieved by \name are particularly remarkable in the SMD and PSM datasets, where it outperforms other baselines by at least 6.8\% and 5.9\% in terms of F1 score and 6.21\% and 2.19\% in terms of R-AUC-PR, respectively. These two datasets exhibit small distribution deviations between anomalous and normal data \cite{tuli2022tranad}, and \name's unconditional imputation design effectively amplifies the gap in imputed error between normal and abnormal data, contributing to its superior performance. Furthermore, \name consistently outperforms other baselines with low F1-std. in the SWaT, SMAP, and GCP datasets, which demonstrates its remarkable robustness. \rv{Interestingly, we observe that all  approaches demonstrate comparatively lower performance in the SwaT dataset. This can be attributed to the intricate and diverse MTS patterns present in the SWaT dataset, underscored by the dataset's expansive training set size and high dimensionality (51).
This leads to challenges in achieving accurate modeling, consequently resulting in inferior anomaly detection performance across all approaches.}

\begin{table}[t]
\caption{The ADD (mean$\pm$std.) performance comparison for all approaches. Results are averaged on 6 runs.\label{tab:add}}
\vspace*{-1.em}
\resizebox{\columnwidth}{!}{
\addtolength{\tabcolsep}{-3pt}
\begin{tabular}{lccccccc}
\hline
Method          & SMD                 & PSM                 & SMAP                 & MSL                  & SWaT                  & GCP                  & Average               \\ \hline
IsolationForest & 90 $\pm$ 1          & 191 $\pm$ 17        & 394 $\pm$ 93         & 123 $\pm$ 28         & 539 $\pm$ 20          & 203 $\pm$ 3          & 257 $\pm$ 27          \\
BeatGAN         & 38 $\pm$ 2          & 166 $\pm$ 11        & 345 $\pm$ 23         & 68 $\pm$ 24          & 607 $\pm$ 6           & 130 $\pm$ 13         & 226 $\pm$ 13          \\
LSTM-AD         & 87 $\pm$ 1          & 224 $\pm$ 54        & 541 $\pm$ 51         & 115 $\pm$ 29         & 627 $\pm$ 4           & 107 $\pm$ 1          & 284 $\pm$ 23          \\
InterFusion     & \textbf{22 $\pm$ 2} & 40 $\pm$ 10         & 423 $\pm$ 4          & \textbf{32 $\pm$ 15} & 454 $\pm$ 141         & 141 $\pm$ 1          & 185 $\pm$ 29          \\
OmniAnomaly     & 26 $\pm$ 1          & 121 $\pm$ 11        & 116 $\pm$ 38         & 93 $\pm$ 2           & 550 $\pm$ 48          & 131 $\pm$ 6          & 173 $\pm$ 18          \\
GDN             & 38 $\pm$ 1          & 148 $\pm$ 0         & 402 $\pm$ 4          & 106 $\pm$ 2          & 1478 $\pm$ 0          & 125 $\pm$ 0          & 383 $\pm$ 1           \\
MAD-GAN         & 59 $\pm$ 57         & 122 $\pm$ 2         & 404 $\pm$ 20         & 88 $\pm$ 0           & 926 $\pm$ 337         & 157  $\pm$ 0         & 293 $\pm$ 69          \\
MTAD-GAT        & 90 $\pm$ 100        & 182 $\pm$ 0         & 542 $\pm$ 2          & 96 $\pm$ 17          & 482 $\pm$ 80          & 145  $\pm$  0        & 256 $\pm$ 33          \\
MSCRED          & 32 $\pm$ 0          & 218 $\pm$ 35        & 622 $\pm$ 48         & 109 $\pm$ 30         & 1065  $\pm$ 339       & 145 $\pm$ 3          & 365 $\pm$ 76          \\
TranAD          & 25 $\pm$ 0          & 127 $\pm$ 4         & 291 $\pm$ 2          & 56 $\pm$ 12          & 657 $\pm$ 246         & 104 $\pm$ 11         & 210 $\pm$ 46          \\
\name           & 24 $\pm$ 1          & \textbf{28 $\pm$ 1} & \textbf{98 $\pm$ 31} & 46 $\pm$ 4           & \textbf{350 $\pm$ 43} & \textbf{75 $\pm$  1} & \textbf{104 $\pm$ 14} \\ \hline
\end{tabular}}
\end{table}

\vspace*{-0.5em}
\subsubsection{Timeliness Performance}
In Table~\ref{tab:add}, we present the ADD (mean$\pm$std.) performance comparison on all datasets over 6 runs, along with their average values. 
Observe that \name demonstrates remarkable performance in this aspect as well. Overall, \name achieves the lowest average ADD values (104) with low variance, surpassing other baselines by at least 39.9\%. Upon closer examination of dataset-specific performance, it is observed that \name consistently outperforms other baselines in 4 out of 6 datasets. This indicates that \name is highly sensitive to abnormal points and can capture them at the earliest detection timing. This superior performance can be attributed to the grating masking design employed by \name, which enables to partially envision the future values of the time series in the masked regions. A more detailed ablation analysis on the masking strategy is presented in Sec.~\ref{sec:mask_ablation}. The ADD metric holds significant importance in industrial practice, as early detection of anomalies allows for prompt mitigation of failures \cite{ghosh2022fight}, potentially preventing more severe consequences. The advantage of \name in achieving faster anomaly detection makes it a suitable choice for real-world deployment in systems with high reliability requirements.

\vspace*{-0.5em}
\subsection{Ablation Analysis (RQ2, RQ3)}
\begin{table*}[t]
\caption{Performance comparison on 6 benchmark datasets for all ablation analysis considered in this paper. \label{tab:ablation}}
\vspace*{-1.em}
\resizebox{1\textwidth}{!}{
\begin{tabular}{lccccccccccccccc}
\hline
\multirow{2}{*}{Method} & \multicolumn{5}{c}{SMD}                  & \multicolumn{5}{c}{PSM}                  & \multicolumn{5}{c}{SWaT}                 \\ \cline{2-16} 
                        & P     & R     & F1    & R-AUC-PR & ADD   & P     & R     & F1    & R-AUC-PR & ADD   & P     & R     & F1    & R-AUC-PR & ADD   \\ \hline
\name                   & 0.952 & 0.951 & 0.949 & 0.382    & 23.7  & 0.981 & 0.975 & 0.978 & 0.471    & 28.4  & 0.899 & 0.846 & 0.871 & 0.194    & 350.4 \\
Forecasting             & 0.892 & 0.918 & 0.896 & 0.268    & 22.6  & 0.974 & 0.868 & 0.914 & 0.411    & 96.2  & 0.895 & 0.792 & 0.839 & 0.290    & 451.7 \\
Reconstruction          & 0.641 & 0.796 & 0.682 & 0.106    & 14.1  & 0.891 & 0.898 & 0.894 & 0.291    & 58.3  & 1.000 & 0.657 & 0.793 & 0.564    & 663.1 \\ \hline
Non-ensemble            & 0.934 & 0.953 & 0.941 & 0.230    & 24.5  & 0.975 & 0.974 & 0.974 & 0.390    & 30.6  & 0.898 & 0.831 & 0.861 & 0.248    & 430.5 \\ \hline
Conditional             & 0.955 & 0.951 & 0.951 & 0.395    & 22.9  & 0.977 & 0.962 & 0.969 & 0.425    & 38.4  & 0.932 & 0.849 & 0.888 & 0.217    & 307.0 \\ \hline
Random Mask             & 0.953 & 0.946 & 0.946 & 0.106    & 23.1  & 0.976 & 0.977 & 0.977 & 0.291    & 23.7  & 0.906 & 0.872 & 0.889 & 0.113    & 295.2 \\ \hline\
\rv{w/o spatial transformer}       & \rv{0.951} & \rv{0.947} & \rv{0.946} & \rv{0.337} & \rv{22.2}  & \rv{0.980} & \rv{0.963} & \rv{0.971} & \rv{0.467}  & \rv{31.2}  & \rv{0.940} & \rv{0.895} & \rv{0.867} & \rv{0.353}  & \rv{405.0} \\ 
\rv{w/o temporal transformer}              & \rv{0.899} & \rv{0.892} & \rv{0.884} &  \rv{0.281}  & \rv{28.6}  & \rv{0.964} & \rv{0.973} & \rv{0.969} & \rv{0.391} & \rv{25.8}  & \rv{0.951} & \rv{0.787} & \rv{0.861} & \rv{0.396} & \rv{375.5} \\ \hline\hline
\multirow{2}{*}{Method} & \multicolumn{5}{c}{SMAP}                 & \multicolumn{5}{c}{MSL}                  & \multicolumn{5}{c}{GCP}                  \\ \cline{2-16} 
                        & P     & R     & F1    & R-AUC-PR & ADD   & P     & R     & F1    & R-AUC-PR & ADD   & P     & R     & F1    & R-AUC-PR & ADD   \\ \hline
\name                   & 0.877 & 0.962 & 0.917 & 0.110    & 98.4  & 0.89  & 0.864 & 0.878 & 0.238    & 46.3  & 0.977 & 0.983 & 0.977 & 0.396    & 75.9  \\
Forecasting             & 0.872 & 0.946 & 0.907 & 0.113    & 130.6 & 0.873 & 0.818 & 0.843 & 0.222    & 64.3  & 0.979 & 0.984 & 0.980 & 0.381    & 78.1  \\
Reconstruction          & 0.879 & 0.978 & 0.926 & 0.120    & 72.6  & 0.783 & 0.662 & 0.707 & 0.168    & 105.3 & 0.942 & 0.968 & 0.952 & 0.281    & 59.9  \\ \hline
Non-ensemble            & 0.871 & 0.963 & 0.915 & 0.127    & 99.0  & 0.879 & 0.862 & 0.870 & 0.187    & 57.8  & 0.956 & 0.981 & 0.966 & 0.246    & 86.3  \\ \hline
Conditional             & 0.851 & 0.740 & 0.787 & 0.106    & 287.4 & 0.872 & 0.865 & 0.868 & 0.271    & 52.4  & 0.979 & 0.978 & 0.976 & 0.402    & 81.7  \\ \hline
Random Mask             & 0.920 & 0.908 & 0.913 & 0.180    & 293.5 & 0.888 & 0.896 & 0.892 & 0.168    & 49.1  & 0.975 & 0.980 & 0.975 & 0.406    & 78.8  \\ \hline
\rv{w/o spatial transformer}             & \rv{0.816} & \rv{0.579} & \rv{0.677} & \rv{0.107} & \rv{360.0}  & \rv{0.865} & \rv{0.889} & \rv{0.876} & \rv{0.243} & \rv{48.3}  & \rv{0.981} & \rv{0.925} & \rv{0.938} &   \rv{0.462}  & \rv{99.8} \\ 
\rv{w/o temporal transformer}             & \rv{0.873} & \rv{0.964} & \rv{0.916} &   \rv{0.110}  & \rv{108.1}  & \rv{0.873} & \rv{0.863} & \rv{0.867} &  \rv{0.168} & \rv{43.3}  & \rv{0.976} & \rv{0.998} & \rv{0.986} & \rv{0.398}  & \rv{67.4} \\ \hline
\end{tabular}}
\vspace*{-1.em}
\end{table*}

\begin{table}[t]
\caption{Average results over all datasets of ablation analysis.\label{tab:ablation_avg}}
\vspace*{-1.em}
\label{Avearge results for all ablation study}
\resizebox{1\columnwidth}{!}{
\begin{tabular}{lccccc}
\hline
Method         & P               & R               & F1              & R-AUC-PR        & ADD          \\ \hline
\name    & 0.9298          & 0.9301 & 0.9284          & 0.2986          & 104 \\
Forecasting    & 0.9139          & 0.8876          & 0.8966          & 0.2808          & 141          \\
Reconstruction & 0.8559          & 0.8266          & 0.8256          & 0.2550          & 162          \\ \hline
Non-ensemble   & 0.9187          & 0.9273          & 0.9211          & 0.2380          & 121          \\ \hline
Conditional    & 0.9278          & 0.8910          & 0.9066          & 0.3026 & 132          \\ \hline
Random Mask    & 0.9363 & 0.9298          & 0.9318 & 0.2107          & 127          \\ \hline
\rv{w/o spatial transformer}    & \rv{0.9224}        & \rv{0.8514}          & \rv{0.8794}         &  \rv{0.3280} & \rv{161}          \\ 
\rv{w/o temporal transformer}    & \rv{0.9229} & \rv{0.9131}          & \rv{0.9139} &       \rv{0.2910}  & \rv{108}          \\ \hline
\end{tabular}}
\end{table}

Next, we conduct a comprehensive ablation analysis to evaluate the effectiveness of each design choice in \name, shedding light on how these design choices contribute to enhancing the anomaly detection performance. The aggregated results specific to each dataset are presented in Table~\ref{tab:ablation}, while Table~\ref{tab:ablation_avg} showcases the average results across all datasets. The reported results are the average of 6 independent runs. Note that in the tables, ``\name'' \emph{represents the combination of the following designs: Imputation, Ensembling, Unconditional, Grating Masking \rv{and full \model}.}

\subsubsection{Imputation vs. Forecasting vs. Reconstruction\label{sec:unsupervised_ablation}}
First, we compare the anomaly detection performance of different MTS modeling approaches, namely imputation, forecasting, and reconstruction. In the case of forecasting and reconstruction, we adopt the same configuration as \name, with the only distinction being the forecasting method that predicts future values given historical observations, while the reconstruction method corrupts all values with noise vectors and reconstructs them. Overall, the \name framework, which utilizes the imputation method, achieves the highest performance in terms of accuracy and timeliness on average, outperforming other MTS modeling methods by at least 3.18\% on F1 score, 1.78\% on R-AUC-PR and 26.2\% ADD. In addition, we observe that forecasting outperforms reconstruction, indicating that incorporating historical information leads to improved performance of the self-supervised model. Moreover, as shown in Table~\ref{tab:ablation}, \name achieves the highest F1 score and ADD in 5 out of 6 datasets, and the highest R-AUC-PR on 4 out of 6 datasets, highlighting the accuracy and robustness of the imputation approach.

\begin{figure}[t]
\centering
\includegraphics[width=1\columnwidth]{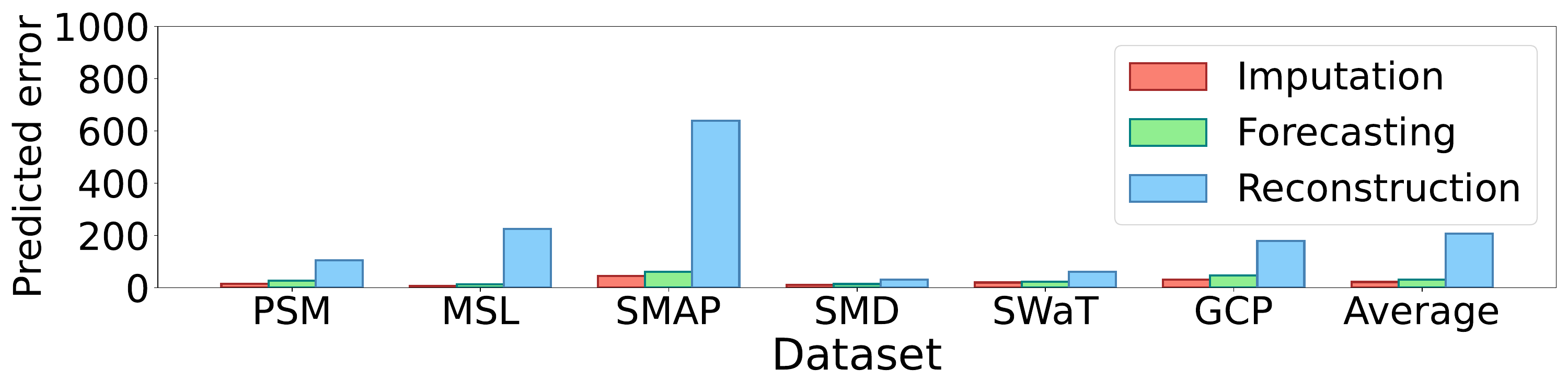}
\vspace*{-2.5em}
\caption{Predicted error of imputation, forecasting and reconstruction approaches on all datasets.
\label{fig:pred_e}}
\end{figure}


The performance improvement can be attributed to the superior self-supervised modeling quality achieved through the imputation approach. Fig.~\ref{fig:pred_e} illustrates the predicted error of each modeling approach, along with their average values. A lower prediction error signifies a more accurate modeling of MTS, which in turn enhances the performance of anomaly detection. Notably, the imputation approach consistently exhibits the lowest predicted error across all datasets, significantly outperforming the forecasting and reconstruction approaches. These results indicate the imputation approach's superior self-supervised modeling capability. They further validate that enhancing the self-supervised modeling ability contributes to
the anomaly detection performance for MTS data.

\subsubsection{Ensembling vs. Non-ensembling\label{sec:ensemble_ablation}}
\begin{figure}[t]
\centering
\vspace*{-1.2em}
\includegraphics[width=1\columnwidth]{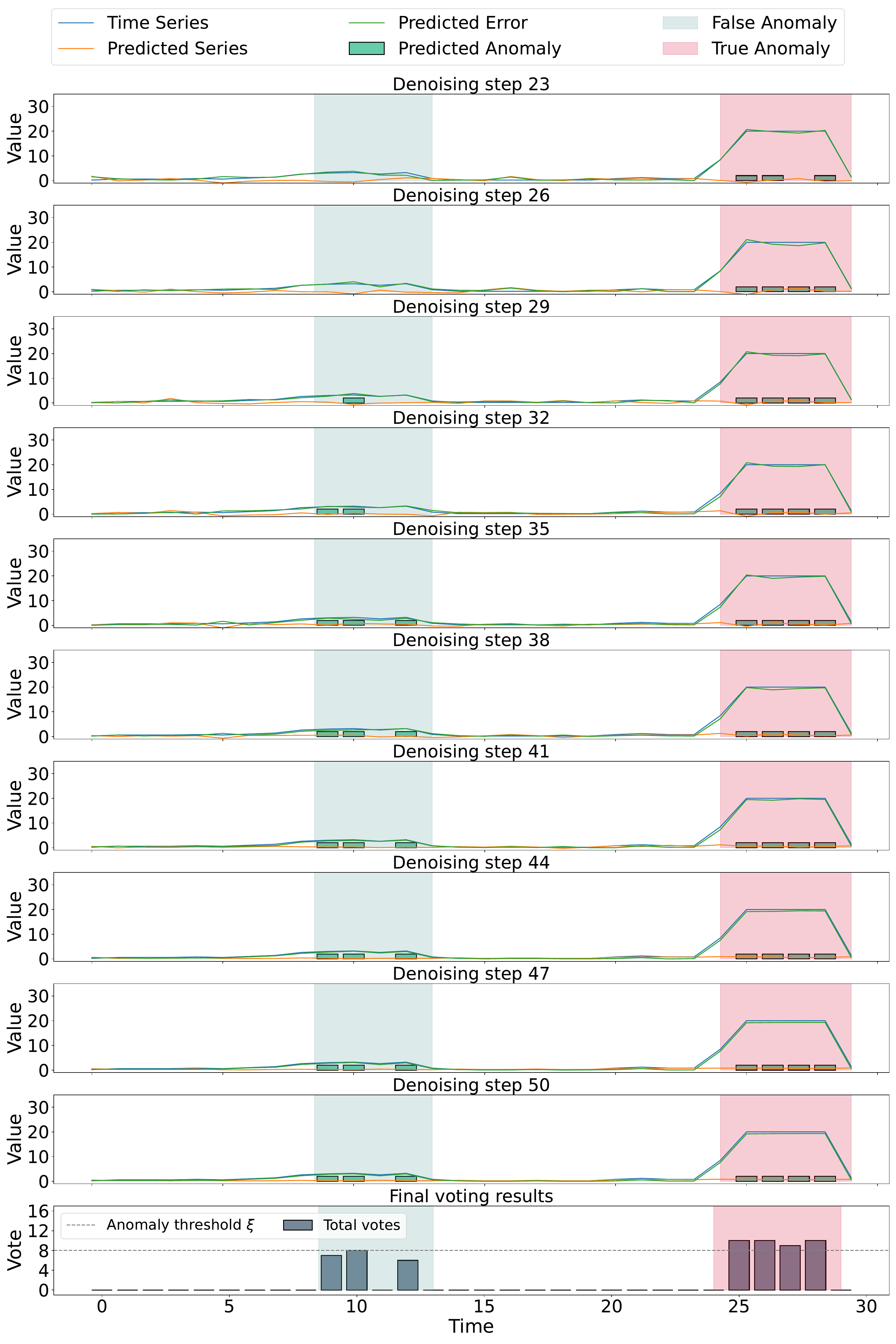}
\vspace*{-2.8em}
\caption{An example of the ensemble inference, showcasing the step-wise prediction and the final voting mechanism.
\label{fig:ensemble_case}}
\end{figure}

Next, we investigate the impact of the ensembling voting mechanism on the anomaly detection performance. The non-ensembling approach solely relies on the final denoised results and applies thresholding on the imputed error for anomaly detection. In comparison, \name on average achieves a 0.73\% higher F1 score, 6.06\% higher R-AUC-PR and a lower 35.8\% ADD compared to the non-ensembling approaches. This suggests that the utilization of the ensembling approach enhances both the accuracy and timeliness performance of anomaly detection, particularly for ranged anomalies. Furthermore, \name consistently outperforms its counterpart across all datasets. Upon closer examination of Table~\ref{tab:ablation_avg}, we observe that \name exhibits a greater advantage in terms of precision over recall. This indicates that the anomalies detected through ensembling are more likely to be true anomalies, thereby reducing the false positive rate.

Fig.~\ref{fig:ensemble_case} illustrates an example on the SMD dataset, demonstrating how the ensembling mechanism improves the anomaly detection performance. The first 10 subplots depict the time series, imputation prediction, predicted error, and anomaly prediction for each of the 10 denoising steps used in the ensembling voting. The final subplot showcases the aggregated voting results for anomalies. Several key insights can be derived from the figure. Firstly, the imputation results progressively improve with each denoising step, aligning with our expectations as diffusion models perform step-by-step imputation. Secondly, relying solely on the final step can lead to false positive predictions (blue shaded area). However, the ensembling voting mechanism plays a crucial role in correcting these false positives. From step 23 to 32, the false positive data receives fewer votes compared to the true positive region (red shaded area), causing them to fall below the final voting threshold (8 votes) and be eliminated from the ensemble's predicted anomalies. This case study provides a clear illustration of how ensembling can enhance accuracy and robustness, showcasing the unique capabilities offered by diffusion models.

\vspace*{-1em}
\subsubsection{Unconditional vs. Conditional Diffusion Models\label{sec:con_ablation}}
\begin{figure}[t]
\centering
\includegraphics[width=0.95\columnwidth]{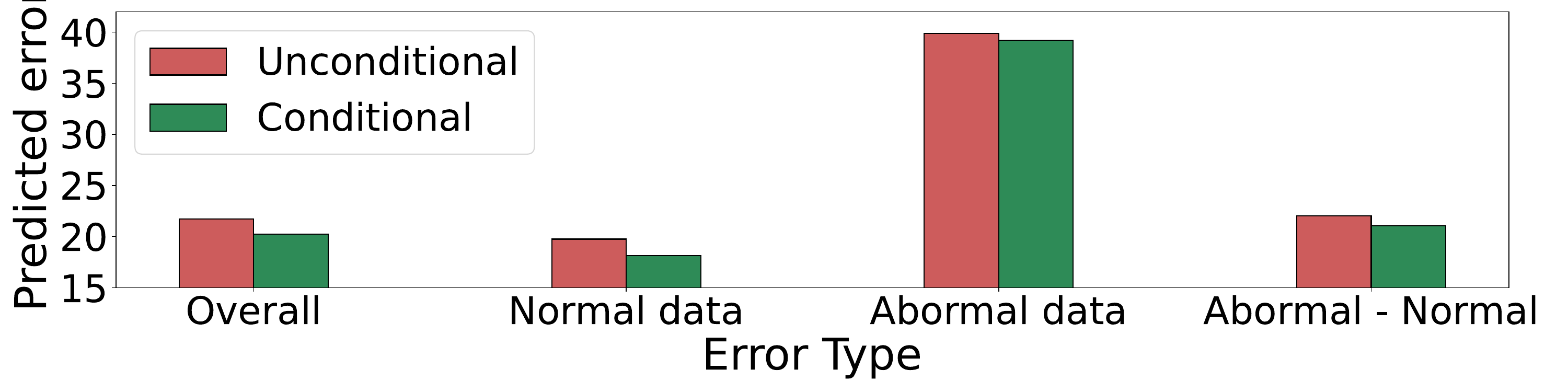}
\vspace*{-1.5em}
\caption{Predicted error of normal/abnormal data comparison on conditional/unconditional diffusion models.
\label{fig:con_e}}
\end{figure}

We now shift our focus to evaluating the effectiveness of the design of unconditional diffusion models described in Sec.~\ref{sec:impdiffusion}. In Table~\ref{tab:ablation_avg}, we observe that \name, which utilizes unconditional diffusion models, achieves superior accuracy and timeliness performance compared to its conditional counterpart, with a 2.1\% higher F1 score and a 21.1\% lower ADD. This gain is particularly pronounced in the SMAP dataset, which comprises shorter time sequences. The R-AUC-PR values obtained from both approaches are comparable, with the conditional method exhibiting a slight advantage.

The improvement can be attributed to the unconditional approach, which expands the predicted error between normal and abnormal data. Fig.~\ref{fig:con_e} showcases the overall predicted error, error on normal data, error on abnormal data, and the difference (abnormal data - normal data) averaged for all datasets. Note that  larger difference in predicted error between abnormal and normal data generally indicates a more distinct classification boundary for thresholding approaches. 
Notably, the unconditional approach generally yields higher overall predicted error. This aligns with our expectations, as the conditional approach makes predictions based on the ground truth time series values, providing more direct and focused guidance compared to the unconditional approach, which predicts solely based on the forward noise in the unmasked regions. However, the error difference between abnormal and normal data is amplified by the unconditional method, as illustrated in the figure. This further confirms the effectiveness of the unconditional approach, as it establishes a clearer error decision boundary for thresholding, enabling better discrimination of anomalies.

\subsubsection{Grating Masking vs. Random Masking\label{sec:mask_ablation}}
We compare the performance of the grating masking and random masking designs as introduced in Sec.~\ref{sec:masking}. Interestingly, on average, the two approaches achieve comparable F1 scores, with random masking slightly outperforming grating masking by 0.4\%. The accuracy performance of the two masking designs is also quite similar across datasets. However, it is worth noting that the grating masking design consistently outperforms the random masking design in terms of R-AUC-PR. On average, the grating masking design achieves an 8.79\% higher score compared to the random masking design, outperforming its counterpart on 4 out of 6 datasets. This result suggests that \name exhibits higher accuracy in detecting ranged anomalies. This is particularly relevant in real-world scenarios where such ranged anomalies occur frequently. In addition, we observe that grating masking exhibits a significant advantage in terms of ADD, with a gain of 18.4\%. This can be attributed to the value envisioning property in the windowed masked regions of the grating design, which is absent in random masking. Therefore, grating masking is more suitable for industrial applications where timely detection of anomalies is crucial to ensure system reliability.

\subsubsection{\rv{Components of \modelt\label{sec:model_ablation}}} 


\rv{Finally, we conduct an ablation analysis to evaluate the impact of removing individual components of \model, specifically the spatial and temporal transformers, on anomaly detection performance. As summarized in Table~\ref{tab:ablation_avg}, the removal of either the spatial or temporal transformer results in a decrease in F1 and ADD performance compared to the complete \name model, albeit to varying degrees. Notably, \name without the spatial transformer exhibits poorer performance than when removing the temporal transformer, emphasizing the importance of capturing inter-metric correlations in MTS anomaly detection. This is particularly evident in the SMAP dataset,  which also exhibits strong interrelations between metrics, as shown in Table~\ref{tab:ablation}. Employing the spatial transformer significantly improve the anomaly detection performance on it.}

\rv{Conversely, the removal of the temporal transformer also leads to a decline in performance across all metrics. As shown in Table \ref{tab:ablation}, the most substantial impact is observed in the SMD dataset, with a significant performance drop compared to other datasets. This emphasizes the importance of accurately modeling and weighting the time dimension in the SMD dataset, where temporal correlations play a crucial role. Consequently, both the spatial and temporal transformers play pivotal roles in enhancing \name's anomaly detection performance, underscoring the superiority of the design of \model.}

\vspace*{-0.5em}
\section{Production Impact and Efficiency}
The proposed \name has been integrated as a critical component within a large-scale email delivery microservice system at Microsoft. This system consists of more than 600 microservices distributed across 100 datacenters worldwide, generating billions of trace data points on a daily basis \cite{wang2023root}. \name serves as a latency monitor for email delivery, for detecting any delay regression in each microservice, which may indicate the occurrence of an incident. The online latency data for each microservice are sampled at a frequency of every 30 seconds. In order to assess the performance of \name, we deployed it online and operated over a period of 4 months. We compared the results obtained by \name with a legacy deep learning-based MTS anomaly detector, which has been in operation for years.

Table~\ref{tab:online} presents the online improvements achieved by \name compared to the legacy detector over a period of 4 months\footnote{To comply with confidentiality requirements, the actual numbers for all metrics are omitted, and only the relative improvements are reported.}. The evaluation of efficiency was conducted on containers equipped with Intel(R) Xeon(R) CPU E5-2640 v4 processors featuring 10 cores. Observe that the replacement of the legacy detector with \name has resulted in significant enhancements in anomaly detection accuracy and timeliness, as evidenced by the substantial improvements across all evaluation metrics. Specifically,  compared to the previous online solution, \name exhibits a performance improvement of 11.4\% in terms of F1 score, 14.4\% improvement in terms of R-AUC-PR, and 30.2\% reduction on ADD.
Despite the requirement of multiple inferences to obtain the final results, the online efficiency of \name remains well within an acceptable range. Considering that the latency data are sampled every 30 seconds, performing inference at a rate of 5.8 data points per second is more than sufficient to meet the online requirements. 

\rv{The reliability assessment of a cloud system encompasses two  aspects: \emph{(i)} detection accuracy and \emph{(ii)} detection timeliness of anomalies or incidents \cite{ghosh2022fight}. The notable performance improvements achieved by \name have made a significant impact on the Microsoft email delivery system from the above perspectives, as they have led to considerable time savings in incident detection (TTD), reduced the number of false alarms triggered by the legacy approach, and ultimately enhanced the system's reliability.}

\begin{table}[t]
\centering
\caption{Online performance of \name in production compared to the legacy detector.\label{tab:online}}
\vspace*{-1em}
\begin{tabular}{cccccc}
\hline
\multicolumn{5}{c}{Improvement}   & \multirow{2}{*}{\begin{tabular}[c]{@{}c@{}}Inference efficiency\\ {[}points/second{]}\end{tabular}} \\ \cline{1-5}
P     & R     & F1    & R-AUC-PR & ADD &                                                                                           \\ \hline
9.0\% & 12.7\% & 11.4\% &   14.4\%       &  30.2\%   & 5.8                                                                                       \\ \hline
\end{tabular}
\end{table}


\section{Conclusion}
This paper presents \name, a novel framework that combines time series imputation and diffusion models to achieve accurate and robust anomaly detection in MTS data. By integrating the imputation method with a grating masking strategy, the proposed approach facilitates more precise self-supervised modeling of the intricate temporal and interweaving correlations that are characteristic of MTS data, which in turn enhances the performance of anomaly detection. Moreover, \name employs dedicated diffusion models for imputation, effectively capturing the stochastic nature of time series data. The framework also leverages multi-step denoising outputs unique to diffusion models to construct an ensemble voting mechanism, further enhancing the accuracy and robustness of anomaly detection. Notably, \name is the first to employ time series imputation for anomaly detection and to utilize diffusion models in this context. Extensive experiments on public datasets demonstrate that \name outperforms state-of-the-art baselines in terms of accuracy and timeliness. Importantly, \name has been deployed in real production environments within Microsoft's email delivery system, serving as a core latency anomaly detector and significantly improving system reliability.

\clearpage 
\balance
\bibliographystyle{ACM-Reference-Format}
\bibliography{reference}

\end{document}